\documentclass[runningheads]{llncs}
\usepackage{graphicx}
\graphicspath{{../images/}}

\usepackage{color}
\usepackage{graphicx} 
\usepackage{tabularx}
\usepackage{amsmath}
\usepackage{amsfonts}
\usepackage{amssymb}
\usepackage{mathtools}
\usepackage{dsfont}
\usepackage{centernot}
\usepackage{setspace}
\usepackage[makeroom]{cancel}
\usepackage[bookmarks]{hyperref}

\hypersetup{
    pdftitle={An introduction to domain adaptation and transfer learning},
    pdfauthor={Wouter M. Kouw, Marco Loog},
    pdfsubject={Machine Learning, Domain Adaptation, Transfer Learning},
    pdfkeywords={Machine Learning,Pattern Recognition,Domain Adaptation,Transfer Learning,Covariate Shift,Sample Selection Bias},
    bookmarksnumbered=true,     
    bookmarksopen=true,         
    bookmarksopenlevel=1,       
    colorlinks=false,            
    pdfstartview=Fit,           
    pdfpagemode=UseOutlines,    
    pdfpagelayout=TwoPageRight
}

\newcommand\given [1][]{\:#1\vert\:}

\begin{document}
\title{TECHNICAL REPORT}
\subtitle{An introduction to\\ domain adaptation and transfer learning}
\titlerunning{An introduction to domain adaptation and transfer learning}
%
\author{Wouter M. Kouw, Marco Loog}
\authorrunning{W.M. Kouw, M. Loog}
%
\institute{Delft University of Technology\\Van Mourik Broekmanweg 6, 2628 XE Delft, the Netherlands\\
\email{wmkouw@gmail.com}}
\maketitle              
\begin{abstract}
In machine learning, if the training data is an unbiased sample of an underlying distribution, then the learned classification function will make accurate predictions for new samples. However, if the training data is \emph{not} an unbiased sample, then there will be differences between how the training data is distributed and how the test data is distributed. Standard classifiers cannot cope with changes in data distributions between training and test phases, and will not perform well. \emph{Domain adaptation} and \emph{transfer learning} are sub-fields within machine learning that are concerned with accounting for these types of changes. Here, we present an introduction to these fields, guided by the question: when and how can a classifier generalize from a source to a target domain? We will start with a brief introduction into risk minimization, and how transfer learning and domain adaptation expand upon this framework. Following that, we discuss three special cases of data set shift, namely prior, covariate and concept shift. For more complex domain shifts, there are a wide variety of approaches. These are categorized into: importance-weighting, subspace mapping, domain-invariant spaces, feature augmentation, minimax estimators and robust algorithms. A number of points will arise, which we will discuss in the last section. We conclude with the remark that many open questions will have to be addressed before transfer learners and domain-adaptive classifiers become practical.
\keywords{Machine Learning \and Pattern Recognition \and Domain Adaptation \and Transfer Learning \and Covariate Shift \and Sample Selection Bias.}
\end{abstract}
\newpage
\section{Introduction}
Intelligent systems learn from data to recognize patterns, predict outcomes and make decisions \cite{jordan2015machine,ghahramani2015probabilistic}. In data-abundant problem settings, such as recognizing objects in images, these systems achieve super-human levels of performance \cite{he2015delving}. Their strength lies in their ability to process large amounts of examples and obtain a detailed estimate of what \emph{does} and \emph{does not} constitute the object of interest. Machine learning has become popular due to large-scale data collection and open access of data sets. Learning algorithms are now incorporated into self-driving cars \cite{wang2013adaptive}, drone guidance \cite{giusti2016machine}, computer-assisted diagnosis \cite{kononenko2001machine}, online commerce \cite{lu2015recommender}, satellite cartography \cite{wieland2014performance}, exo-planet discovery \cite{ball2010data}, and machine translation \cite{zhang2015deep}, among others.

\paragraph{} "Intelligence" refers to a computer's ability to \emph{learn} to perform a task \cite{russell2009artificial}. Supervised systems learn through \emph{training}, where the system is rewarded or punished based on whether it produces the right output for a given input \cite{bishop2006pattern,mohri2012foundations}. In order to train an intelligent system, a set of matching inputs and outputs is required. Most often, inputs consist of complicated objects such as images while outputs consists of decisions such as 'yes' or 'no', or classes such as 'healthy', 'at risk', and 'disease'. The system will consider many classification functions on the set of inputs and select the function that produced the smallest error cost. If the examples in the data set are similar to new inputs, then the system will make accurate decisions in the future as well. Classifying new inputs based on a finite set of examples, is called \emph{generalization}. For example, suppose patients are measured on various biometrics such as blood pressure, and have been classified as 'healthy' or 'disease'. Then, a system can be trained by finding the decision function that produces the best diagnoses. If the patients are an accurate reflection of the population of all possible patients, then the trained system will produce accurate diagnoses for new patients as well.

\paragraph{} However, if the collected data it is \emph{not} an accurate reflection of the population, then the system will \emph{not} generalize well. Data is \emph{biased} if certain events are observed more frequently than usual. For example, data collected from older patients is biased with respect to the total human population. If data is biased, then the system will think that certain outcomes are more likely to occur. For example, it might consider certain levels of blood pressure to be normal, when they would actually indicate a health risk for younger patients. Researchers in statistics and social sciences have long studied problems with sample biases and have developed a number of techniques to correct for biased data \cite{heckman1977sample,heckman1990varieties,helton2006survey}. However, generalizing towards wider populations is perhaps too ambitious. Instead, machine learning researchers are targeting specific other populations. For instance, can we use information from \emph{adult} humans to train an intelligent system for diagnosing \emph{infant} heart disease? 



\paragraph{} Such problem settings are known as \emph{domain adaptation} or \emph{transfer learning} settings \cite{ben2010theory,patel2015visual,pan2010survey}. The population of interest is called the \emph{target domain}, for which labels are usually not available and training a classifier is not possible. However, if data from a similar population is available, it could be used as a source of additional information. Now the challenge is to overcome the differences between the domains so that a classifier trained on the source domain generalizes well to the target domain. Such a method is called a \emph{domain-adaptive classifier} or a \emph{transfer learner} (the difference will be defined in Section \ref{sec:domain}). Generalizing across distributions is difficult and it is not clear which conditions have to be satisfied for a classifier to perform well. We therefore focus on the question: when and how can a statistical classifier generalize from a source to a target domain?

\subsection{Relevance}
A more detailed example of adaptation is the following: in clinical imaging settings, radiologists manually annotate tissues, abnormalities, and pathologies of patients. Biomedical engineers then use these annotations to train systems to perform automatic tissue segmentation or pathology detection in medical images. Now suppose a hospital installs a new MRI scanner. Unfortunately, due to the mechanical configuration, calibration, vendor and acquisition protocol of the scanner, the images it produces will differ from images produced by other scanners \cite{van2015transfer,ghafoorian2017transfer,kouw2017mr}. Consequently, systems trained on data from other scanners would fail to perform well on the new scanner. An adaptive system could find correspondences in images between scanners, and change its decisions accordingly. Thus it avoids the time, funds and energy needed to annotate images from the new scanner \cite{van2015transfer,ghafoorian2017transfer,kouw2019learning}.

\paragraph{} Other examples of domain adaptation and transfer learning in fields that employ machine learning include: in bioinformatics, adaptive approaches have been successful in sequence classification \cite{widmer2010novel,mei2011gene}, gene expression analysis \cite{chen2010new,xu2011multi}, and biological network reconstruction \cite{nassar2008new,kato2012transfer}. Most often, domains correspond to different model organisms or different data-collecting research institutes \cite{xu2011survey}. In predictive maintenance, every time the fault prognosis system raises an alarm and designates that a component has to replaced, the machine changes its properties \cite{caesarendra2010machine}. The system will have to adapt to the new setting, until another component is replaced. In search-and-rescue robotics, a system that needs to autonomously navigate wilderness trails will have to adapt to detect concrete structures if it is to be deployed in an urban environment \cite{giusti2016machine,wieland2014performance}. Computer vision systems that recognize activities have to adapt across different surroundings as well as different groups of people \cite{van2010transferring,hachiya2012importance,gedik2016speaking}. In natural language processing, texts from different publication platforms are tricky to analyze due to different contexts and differences between how authors express themselves. For instance, financial news articles use a vocabulary that differs from the one in biomedical research abstracts \cite{blitzer2006domain}. Similarly, online movie reviews are linguistically different from tweets \cite{peddinti2011domain}. Sentiment classification relies heavily on context as well; different words are used to express whether someone likes a book versus whether an electronic gadget \cite{blitzer2007biographies,he2016ups}. 

\paragraph{} In some situations, the target is a sub-population of the source domain. For instance, in \emph{personalized} systems, the target is a single individual while the source might be a set of individuals. One of the first types of personalized systems are spam filters: they are often initialized using data from many individuals but will adapt to specific users over time \cite{nuruzzaman2011independent}. Male users receive different kinds of spam than female users for instance, which the system can detect based purely on text statistics. Alternatively, in speaker recognition, an initial speaker-independent system can adapt to new speakers \cite{leggetter1995maximum}. Similarly, general face recognition systems can be adapted to specific persons \cite{zen2014unsupervised} and person-independent activity recognition algorithms can be specialized to particular individuals \cite{gedik2017personalised}. 

\subsection{Outline}
This report is guided by the questions: \emph{when} and \emph{how} can a classifier generalize from a source to a target domain? The matter of \emph{when} is discussed through generalization error bounds in Sections \ref{sec:generalization}, \ref{sec:cd-geb} and \ref{sec:iw}, and through types of data shifts in Section \ref{sec:shifts}. The matter of \emph{how} is discussed in Section \ref{sec:approaches}, where we categorize a series of approaches.


\paragraph{} The remainder of the report is structured as follows: Firstly, we will give a brief overview of statistical classification in the following Section. Readers familiar with the material may skip this Section. Secondly, Section \ref{sec:domain} describes how domain adaptation and transfer learning fit within the risk minimization framework. We present stricter definitions of "domains" and "adaptation" as well. Thirdly, Section \ref{sec:shifts} discusses various types of simple data set shifts. Simple in the sense that various aspects of the resulting distributions remain constant between domains and one does not require fully labeled data in each domain. Fourthly, Section \ref{sec:approaches} presents an overview of approaches for more complex cases of domain shifts. We discuss popular algorithms briefly, but bear in mind that this list is not exhaustive. Lastly, Section \ref{sec:discussion} discusses a few ideas and questions that have come to mind while reviewing the literature, and Section \ref{sec:conclusion} draws a few conclusions.

\section{Classification} \label{sec:problem}
This section is a brief overview of classification and risk minimization. For broader overviews, see \cite{friedman2001elements,loog2018supervised}. Readers familiar with this material may skip to Section \ref{sec:domain}.

\subsection{Risk minimization}
\begin{figure}[htb]
\includegraphics[width=.48\textwidth]{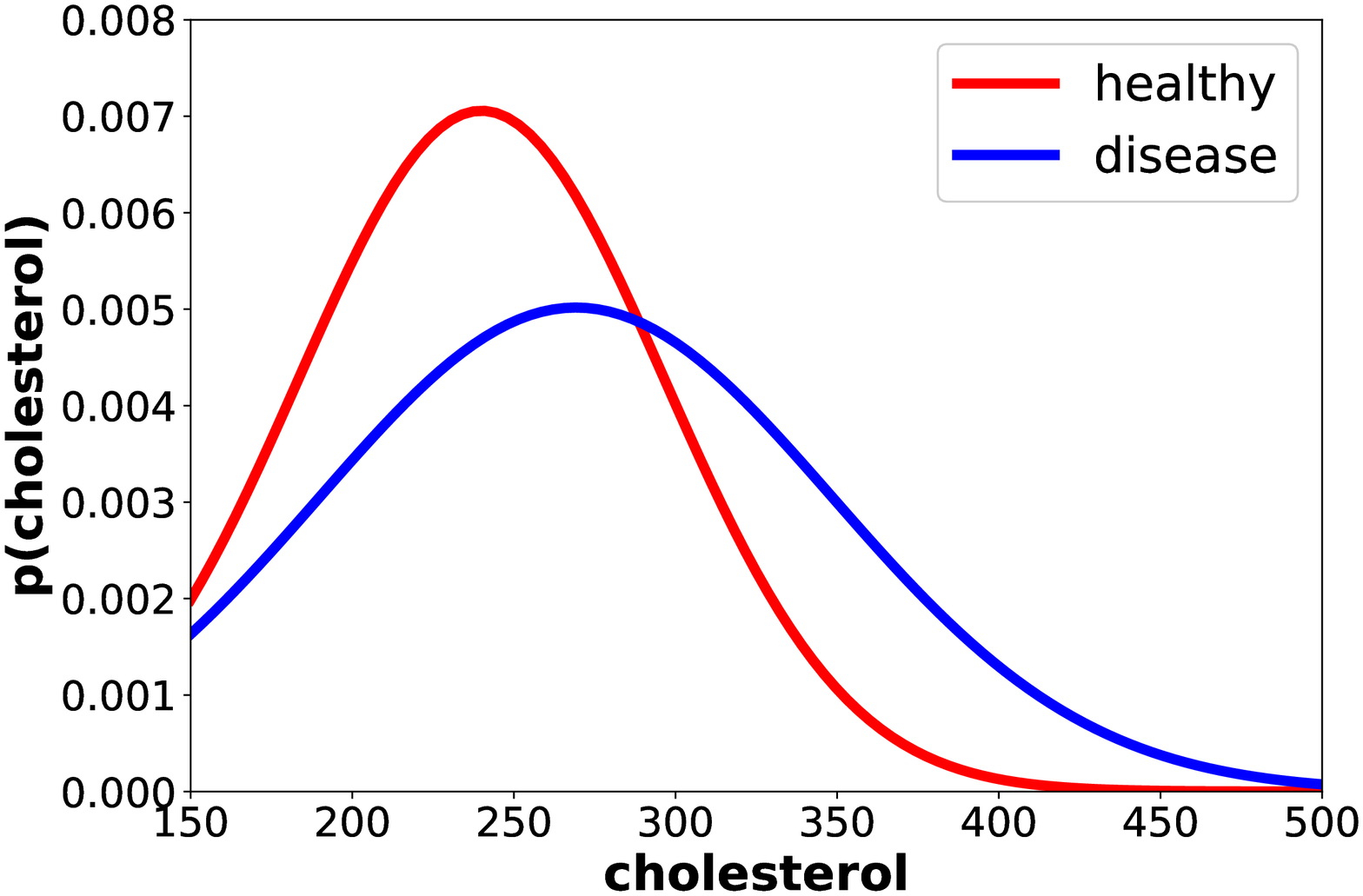} \ 
\includegraphics[width=.48\textwidth]{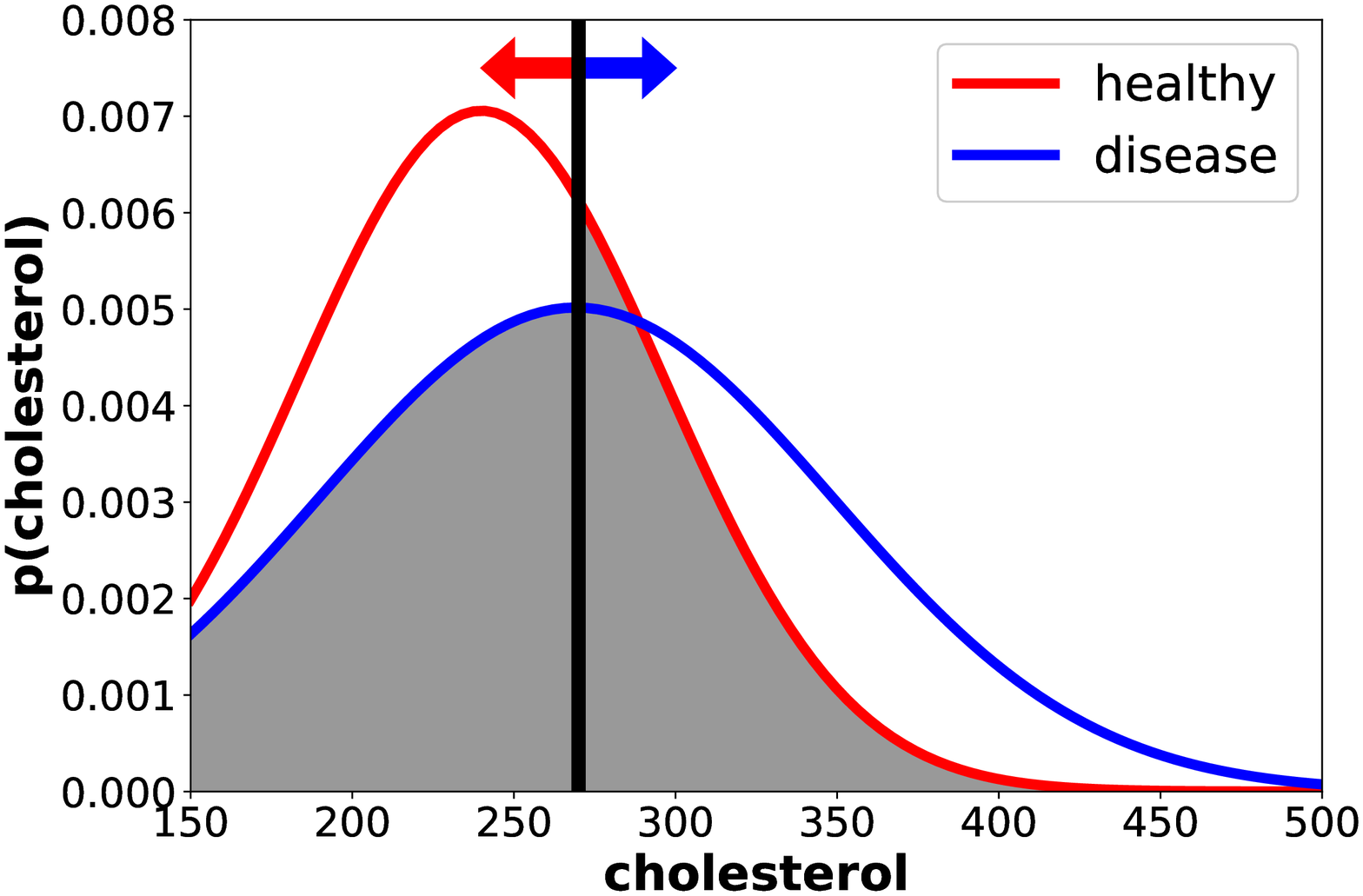}
\caption{Example of a classification problem. (Left) Probability distributions of healthy and ill patients, as a function of their levels of cholesterol. (Right) Classifier, with the black line indicating its decision boundary, and the error consisting of the gray shaded area under the distributions.}
\label{fig:ex_clf}
\end{figure}

One of the most well-known frameworks for the design, construction and analysis of intelligent systems is \emph{risk minimization} \cite{friedman2001elements,mohri2012foundations}. In order to represent an object digitally, we measure one or more \emph{features}. For example, we measure the level of cholesterol of patients in a hospital study. A feature captures information about the object; some levels occur more often than others. These variations over cholesterol $x$ can be described by a probability distribution $p(x)$. Now, suppose we would like to predict whether these patients will develop heart disease. In order to decide between the prognosis 'healthy' and the prognosis 'disease', the system must determine which of the two is more probable for a given level of cholesterol, i.e. $p(\text{healthy} \given x) > p(\text{disease} \given x)$ or $p(\text{healthy} \given x) < p(\text{disease} \given x)$ \cite{theodoridis2010introduction}. Figure \ref{fig:ex_clf} (left) describes two probability distributions as a function of cholesterol level; the red distribution corresponds to 'healthy' and the blue to 'disease'. 

A decision-making problem can be abstractly described as the task of assigning a class, from a finite set of possible classes, to every possible variation of an object. Decision-making systems are therefore called statistical \emph{classifiers}. In their most basic form they consist purely of a function that takes as input an object, encoded by features, and outputs one of the possible classes, e.g. $h(x) = \text{healthy}$. Its output is called its prediction, as there are problem settings where classification errors are unavoidable. We will refer to the classifier itself as $h$, while its prediction is denoted by its application to a particular object $h(x)$. Returning to the heart disease prognosis problem, a classification function for a $1$-dimensional problem can be seen as a threshold, illustrated in Figure \ref{fig:ex_clf} (right) by the black vertical line. It designates everything to the left as 'healthy' and everything to the right as 'disease'. Hence, all heart disease patients left of the line and all healthy patients to the right are misclassified. Note that the current threshold need not be optimal. The classification error is visualized as the gray region under the distributions and can be written as:
\begin{align}
	e(h) =& \ \int_{\mathcal{ X}} [h(x) \neq \text{healthy}] \ p(x \given \text{healthy}) \ p(\text{disease}) \ \mathrm{d}x \nonumber \\
	+& \ \int_{\mathcal{ X}} [h(x) \neq \text{disease}] \ p(x \given \text{disease}) \ p(\text{disease}) \ \mathrm{d}x \, , \nonumber
\end{align}
where $h(x)$ refers to the decision made by the classifier. $p(\text{healthy})$ and $p(\text{disease})$ refer to the probability of encountering healthy patients and heart disease patients in general, while $p(x \given \text{healthy})$ and $p(x \given \text{disease})$ refer to the probabilities of observing a particular level of cholesterol $x$ given that the patient is healthy or has a disease (also known as the \emph{class-conditional} distributions), respectively. 

\paragraph{} The classifier should be able to make a decision over all possible cholesterol levels $\mathcal{ X}$. Since cholesterol is a continuous variable, the decision function is integrated over all possible levels. If the objects were measured on a discrete variable, then the integration would be equivalent to a sum. Essentially, the first term describes how often the classifier will make a mistake in the form of deciding that an actual healthy patient will develop heart disease and the second term describes how often it thinks that a patient with heart disease will be healthy. Summing these two terms constitutes the overall classification error $e(h)$. 

\paragraph{} If 'healthy' and 'disease' are encoded as a variable $y$, then the classification error can be written as follows:
\begin{align}
	e(h) = \sum_{y \in \mathcal{ Y}} \int_{\mathcal{ X}} \left[h(x) \neq y \right] p(x,y) \ \mathrm{d} x \, . \nonumber
\end{align}
where $p(x,y) = p(x \given y) p(y)$. $\mathcal{Y}$ numerically represents the set of classes, in this case $\mathcal{Y} = \{\text{healthy} = -1, \text{disease} = +1\}$. Objects are often not described by one feature but by multiple measured properties. As such, $x$ is a $D$-dimensional random vector $\mathcal{X} \subseteq \mathbb{R}^{D}$.

\subsubsection{Loss functions}
The notion of disagreement between the predicted and the true class can be described in a more general form by using a function that describes the numerical cost of correct versus incorrect classification. This function is known as a \emph{loss} function $\ell$, which takes as input the classifier's prediction for a certain object $h(x)$ and the object's true class $y$. The classification error, $e(h)$, is also known as the \emph{$0$/$1$ loss}, denoted $\ell_{0/1}$. It has value $0$ whenever the prediction is equal to the true label and value $1$ whenever they are not equal; $\ell_{0/1}(h(x),y) = \left[h(x) \neq y \right]$. However, this function is hard to work with; the function has a constant and discontinuous derivative, which means that gradient-based methods cannot be applied. One would have to evaluate each classifier separately in order to find the best one. Most often, there are infinitely many classifiers to consider, which implies infinitely many evaluations. 

\paragraph{} Other loss functions are the \emph{quadratic} / \emph{squared} loss, $\ell_{\text{qd}} (h(x),y) = (h(x) - y)^2$, the \emph{logistic} loss $\ell_{\log}(h(x),y) = y h(x) - \log \sum_{y' \in {\cal Y}} \exp(y'h(x))$ or the \emph{hinge} loss $\ell_{\text{hinge}}(h(x),y) = \max(0, 1 - y h(x))$. These are called \emph{convex surrogate} losses, as they approximate the $0$/$1$ loss through a convex function \cite{bartlett2006convexity}. Convex loss functions are easier to optimize, but do not necessarily lead to the same solution than if the error function was optimized directly. Overall, the choice of a loss function can have a major impact on the behaviour of the resulting classifier.

\paragraph{} Considering that we are integrating the loss function with respect to probabilities, we are actually looking at the expected loss, also called the \emph{risk}, of a particular classifier:
\begin{align}
	R(h) = \mathbb{E}_{\mathcal{ X,Y}} \ \big[ \ell( h(x),y) \big] \, , \label{eq:risk}
\end{align}
where $\mathbb{E}$ stands for the expectation. Its subscript denotes which variables are being integrated over. Given a risk function, we can evaluate multiple possible classifiers and select the one for which the risk is as small as possible:
\begin{align}
	h^{*} = \underset{h}{\arg \min} \ \mathbb{E}_{\mathcal{ X,Y}} \ \big[ \ell( h(x), y) \big] \, . \nonumber
\end{align}
The asterisk superscript denotes optimality with respect to the chosen loss function. There are many ways to perform this minimization step, with vastly different computational costs. The main advantage of convex loss functions is that efficient optimization procedures can be used \cite{boyd2004convex}. 

\subsubsection{Empirical risk}
Up to this point, we have only considered the case where the probability distributions are completely known. In practice, this is rarely the case: only a finite amount of data can be collected. Measurements of objects can be described as a data set $\mathcal{ D}^{n} = \{(x_i,y_i)\}_{i=1}^{n}$, where each $x_i$ is an independent sample from the random variable $\mathcal{ X}$, and is labeled with its corresponding class $y_i$. The expected value with respect to the joint distribution of data and labels can be approximated with the sample average:
\begin{align}
	\hat{R}(h \given \mathcal{D}^{n}) = \frac{1}{n} \sum_{i=1}^{n} \ \ell( h(x_i), y_i) \, . \nonumber
\end{align}
$\hat{R}$ is called the \emph{empirical risk} function. It evaluates classifiers \emph{given} a particular data set. Note that the true risk $R$ from Equation \ref{eq:risk} does not depend on observed data. Minimizing the empirical risk with respect to a classifier for a particular data set, is called \emph{training} the classifier:
\begin{align}
	\hat{h} = \underset{h \in \mathcal{H}}{\arg \min} \ \hat{R}(h \given \mathcal{D}^{n}) \, \, \nonumber
\end{align}
where $\mathcal{H}$ refers to the collection of all possible classifiers that we consider, also known as the hypothesis space. A risk-minimization system is said to \emph{generalize} if it uses information on specific objects to make decisions for all possible objects.

\paragraph{} In general, more samples lead to better approximations of the risk, and the resulting classifier will be closer to the optimal one. For $n$ samples that are independently drawn and identically distributed, due to the law of large numbers, the empirical risk converges to the true risk \cite{wasserman2013all,boucheron2013concentration}. It can be shown that the resulting classifier will converge to the optimal classifier \cite{vapnik2000nature,mohri2012foundations}. The minimizer of the empirical risk deviates from the true risk due to the estimation error, i.e. the difference between the sample average and the actual expected value, and the optimization error, i.e. the difference between the true minimizer and the one obtained through the optimization procedure \cite{bousquet2008tradeoffs,mohri2012foundations}. 

\subsubsection{Generalization} \label{sec:generalization}
 Ultimately, we are not interested in the error of the trained classifier on the given data set, but in the error on all possible future samples $e(h) = \mathbb{E}_{\mathcal{ X,Y}} [h(x) \neq y]$. The difference between the true error and the empirical error is known as the \emph{generalization error}: $e(h) - \hat{e}(h)$ \cite{bartlett2003prediction,mohri2012foundations}. Ideally, we would like to know if the generalization error will be small, i.e., that our classifier will be \emph{approximately correct}. However, because classifiers are functions of data sets, and data sets are random, we can only describe how \emph{probable} it is that our classifier will be approximately correct. We can say that, with probability $1 - \delta$, where $\delta > 0$, the following inequality bolds (Theorem 2.2 from \cite{mohri2012foundations}):
\begin{align}
	e(h) - \hat{e}(h)  \ \leq \ \sqrt{\frac{1}{2n} \Big( \log |\mathcal{H}| + \log \frac{2}{\delta} \Big) } \, . \label{eq:pac}
\end{align}
where $| \mathcal{H} | $ denotes the cardinality of the finite hypothesis space, or the number of classification functions that are being considered \cite{valiant1984theory,kearns1994introduction,mohri2012foundations}. This result is known as a Probably Approximately Correct (PAC) bound. In words, the difference between the true error, $e(h)$, and the empirical error, $\hat{e}(h)$, of a classifier is less than the square root of the logarithm of the size of the hypothesis space $|{\cal H}|$, plus the log of $2$ over $\delta$, normalized by twice the sample size $n$.  In order to achieve a similar result for the case of an infinite hypothesis space (e.g. linear classifiers), a measure of the complexity of the hypothesis space is required.


\paragraph{} Generalization error bounds are interesting because they analyze what a classifier's performance depends on. In this case, it suggest choosing a smaller or simpler hypothesis space when the sample size is low. Many variants of bounds exist. Some use different measures of complexity, such as Rademacher complexity \cite{bartlett2002rademacher} or Vapnik-Chervonenkis dimensions \cite{bendavid1995characterizations,vapnik2000nature}, while others use concepts from Bayesian inference \cite{mcallester2003simplified,langford2003pac,begin2014pac}.

\paragraph{} Bounds can incorporate assumptions on the problem setting \cite{bartlett1998sample,mohri2012foundations,devroye2013probabilistic}. For example, one can assume that the posterior distributions in each domain are equal and obtain a bound for a classifier that exploits that assumption (c.f. Equation \ref{eq:geb_iw}). Assumptions restrict the problem setting, i.e., settings where that assumption is invalid are disregarded.  This often means that the bound is tighter and a more accurate description of the behaviour of the classifier can be found. Such results have inspired new algorithms in the past, such as Adaboost or the Support Vector Machine \cite{freund1995desicion,cortes1995support}. 

\subsubsection{Regularization}
\begin{figure}[htb]
\centering
	\includegraphics[width=.32\textwidth]{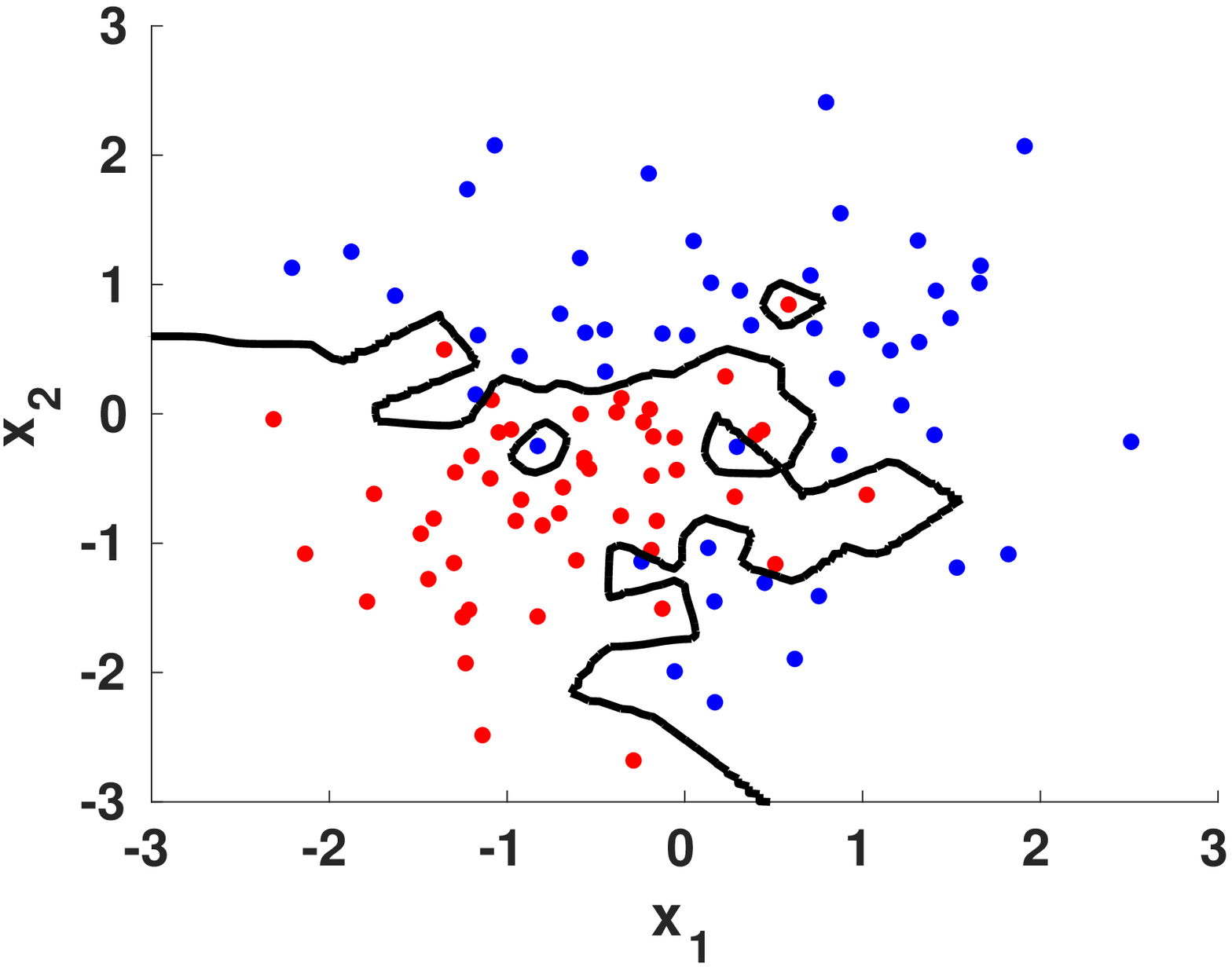} \
	\includegraphics[width=.32\textwidth]{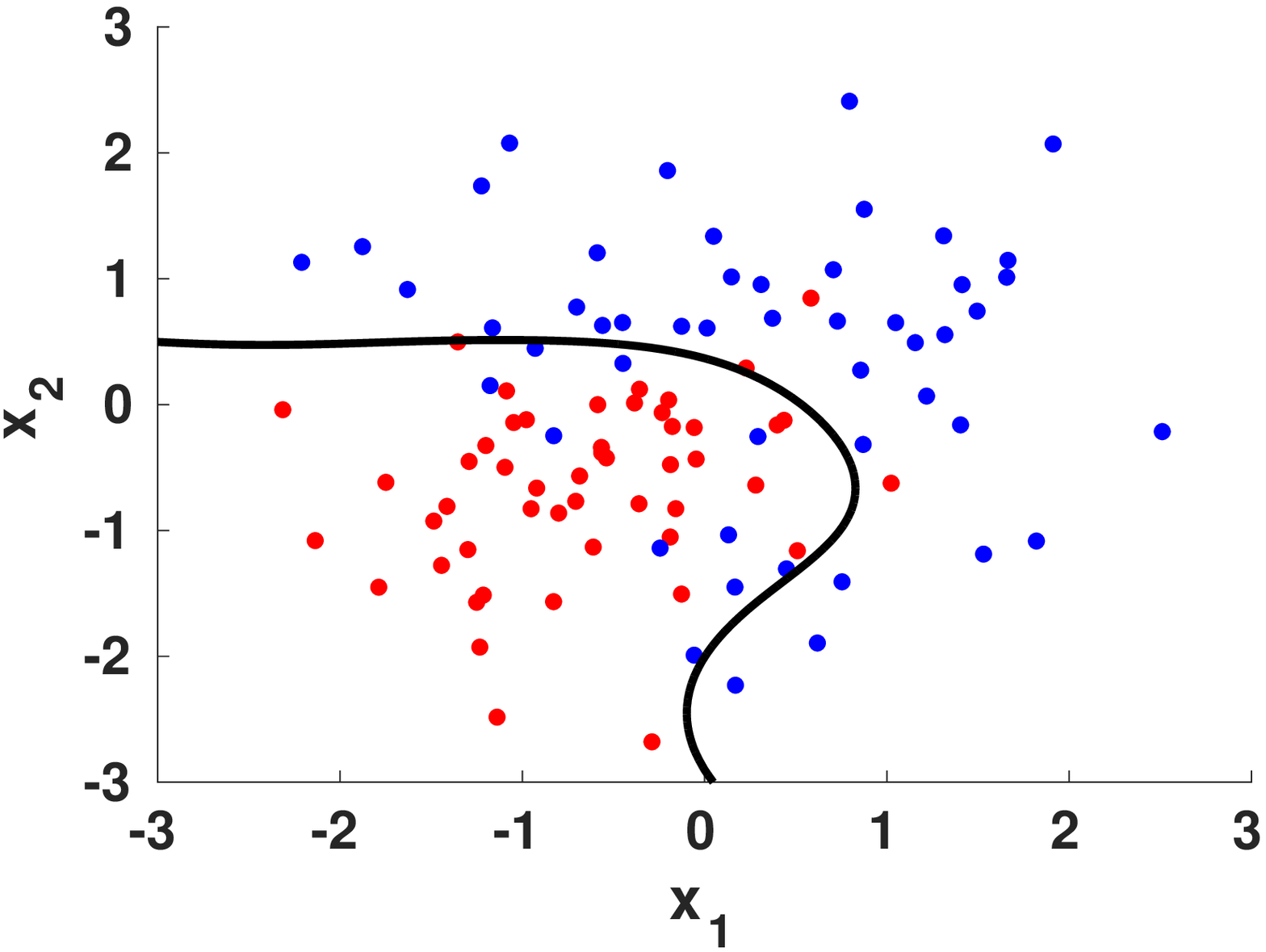} \
	\includegraphics[width=.32\textwidth]{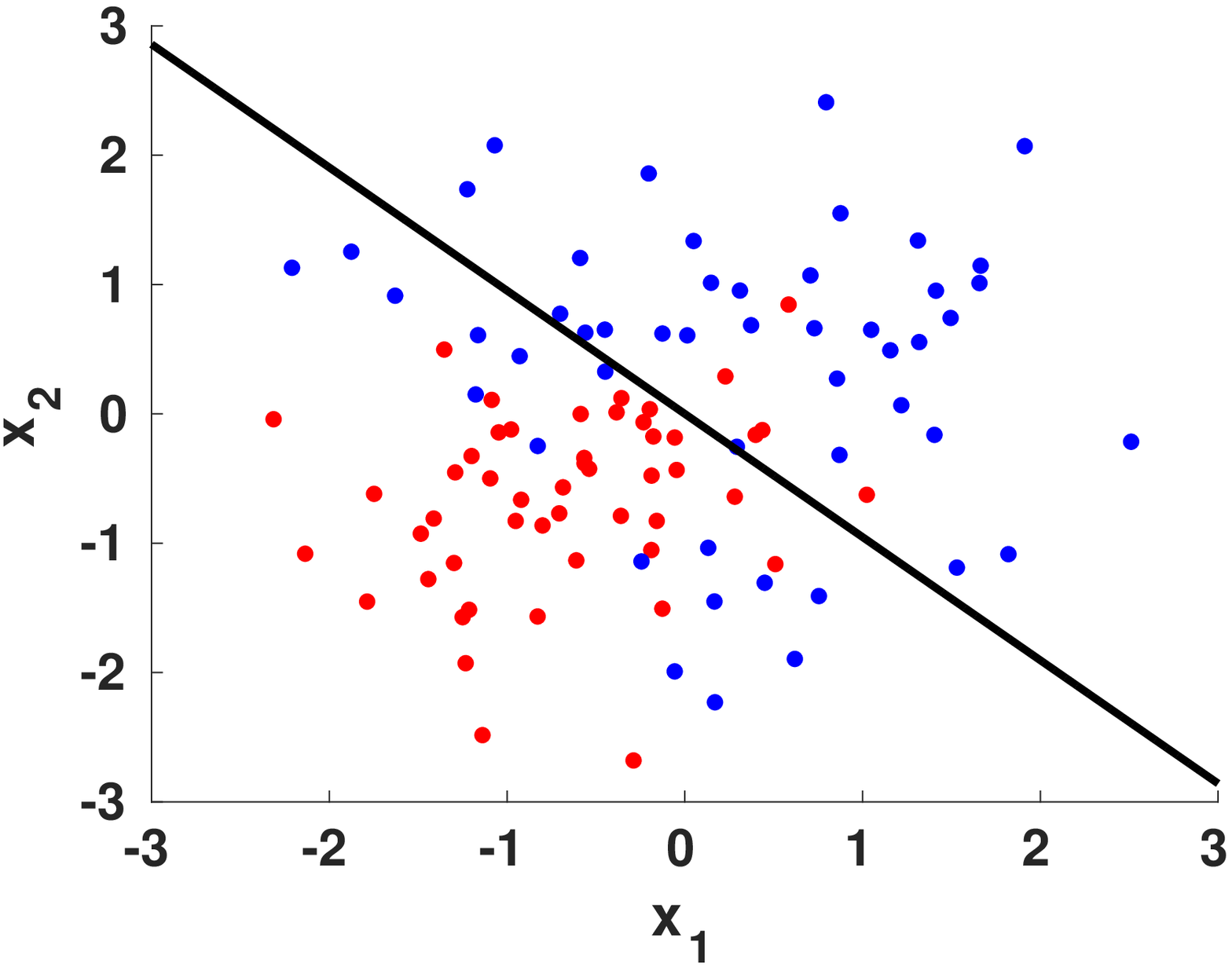}
\caption{Examples of classifier complexities. (Left) Overfitted classifier, (middle) well-fitted classifier, (right) underfitted classifier.}
\label{fig:reg}
\end{figure}

Generalization error bounds tell us that the complexity, or flexibility, of a classifier has to be traded off with the number of available training samples \cite{ehrenfeucht1989general,vapnik2000nature,devroye2013probabilistic}. In particular, a flexible model can minimize the error on a given data set completely, but will be too specific to generalize to new samples. This is known as \emph{overfitting}. Figure \ref{fig:reg} (left) illustrates an example of $2$-dimensional classification problem with a classifier that has perfectly fitted to the training set. As can be imagined, it will not perform as well for new samples. In order to combat overfitting, an additional term is introduced in the empirical risk estimator that punishes model flexibility. This \emph{regularization} term is often a simple additive term in the form of the norm of the classifier's parameters \cite{tikhonov1963solution,bishop2006pattern}. Figure \ref{fig:reg} (middle) visualizes an example of a properly regularized classifier, that will probably generalize well to new samples. Figure \ref{fig:reg} (right) shows an example of a too heavily regularized classifier, also known as an "underfitted" classifier.

\section{Domain adaptation and transfer learning} \label{sec:domain}
We define domains as the combination of an input space $\mathcal{ X}$, an output space $\mathcal{ Y}$ and an associated probability distribution $p$. Inputs are subsets of the $D$-dimensional real space $\mathbb{R}^{D}$ and are sometimes referred to as feature vectors, or points in feature space.  Outputs are classes, which can be binary, in which case $\mathcal{ Y}$ corresponds to $\{-1,+1\}$, or multi-class, in which case $\mathcal{ Y} = \{1, \dots K \}$. Given two domains, we call them different if they are different in at least one of their constituent components, i.e., the input space, the output space, or the probability density function. \emph{Transfer learning} is defined as the general case where the domains are freely allowed to differ in sample space, label space, distribution or all. For example, image caption generators from computer vision generalize from the "image domain" to the "text domain", which would be an example of differences between feature spaces \cite{karpathy2015deep,gopalan2015domain}. \emph{Domain adaptation} is defined as the particular case where the sample and label spaces remain unchanged and only the probability distributions change.

\subsection{Notation} 
We denote the source domain as $(\mathcal{ X}, \mathcal{ Y}, p_{\mathcal{ S}})$ and will sometimes refer to it in shorthand as $\mathcal{ S}$. The target domain is denoted $(\mathcal{ X}, \mathcal{ Y}, p_{\mathcal{ T}})$ with the shorthand $\mathcal{ T}$. Domain-specific functions will be marked with the subscript $\mathcal{ S}$ or $\mathcal{ T}$. For example, $p_{\mathcal{ T}}(x,y)$ for the target joint distribution, $p_{\mathcal{ T}}(x)$ for the target data marginal distribution and $p_{\mathcal{ T}}(x \given y)$ for the target class-conditional distribution. 

Samples from the source domain are denoted with $(x_i,y_i)$, and the source data set is referred to as $\{(x_i,y_i)\}_{i=1}^{n}$. Note that $x$ refers to an element of the input space $\mathcal{ X}$ while $x_i$ refers to a specific observation drawn from the source distribution, $x_i \sim p_{\mathcal{ S}}$. Likewise, samples from the target domain are denoted with $(z_j,u_j)$, with its data set $\{(z_j,u_j)\}_{j=1}^{m}$. All vectors are row vectors, except if otherwise stated. Capital letters denote matrices, for example $X$ as the whole data set of $n$ source samples of $D$-dimensional vectors.

\subsection{Cross-domain generalization error} \label{sec:cd-geb}
Returning to the question: \emph{when} can a classifier generalize from a source to a target domain? The generalization error bound from Section \ref{sec:generalization} describes how much a classifier trained on samples from a distribution will generalize to new samples from that distribution. But it is based on Hoeffding's inequality, which only describes deviations of the empirical risk estimator from its \emph{own} true risk, not deviations from \emph{other} risks \cite{hoeffding1963probability,mohri2012foundations,boucheron2013concentration}. Since Hoeffding's inequality does not hold in a cross-domain setting, the standard generalization error bound does not hold either.

\paragraph{} Nonetheless, it is possible to derive generalization error bounds if more is known on the relationship between $\mathcal{ S}$ and $\mathcal{ T}$ \cite{ando2005framework,blitzer2008learning,cortes2015adaptation,mansour2009domain,ben2010theory,zhang2012generalization,germain2016new}. For example, one of the first error bounds relies on the condition that the ideal hypothesis on both domains has low error \cite{ben2007analysis,ben2010theory}. As will be shown later, the deviation between the target generalization error of a classifier trained in the source domain $e_{\mathcal{ T}}(\hat{h}_{\mathcal{ S}})$ and the target generalization error of the optimal target classifier $e_{\mathcal{ T}}(h^{*}_{\mathcal{ T}})$ depends on this joint error. If it is too large, then the source trained classifier can never be approximately correct in the target domain. 

\paragraph{} Additionally, we need some measure of how much two domains differ from each other. For this bound, the \emph{symmetric difference hypothesis} divergence ($\mathcal{ H}\Delta \mathcal{ H}$-divergence) is used, which takes two classifiers and looks at to what extent they disagree with each other on both domains \cite{ben2010theory}:
\begin{align}
	d_{\mathcal{ H} \Delta \mathcal{ H}} (p_{\mathcal{ S}}, p_{\mathcal{ T}}) = \ 2 \ \underset{h,h' \in \mathcal{ H}}{\sup} \ \left|  \ \text{Pr}_{\mathcal{ S}} \left[h \neq h' \right] - \text{Pr}_{\mathcal{ T}} \left[h \neq h' \right] \ \right|  \, , \nonumber
\end{align}
where the probability $\text{Pr}$ can be computed through integration: $\text{Pr}_{\mathcal{ S}}[h \neq h'] = \int_{\mathcal{ X}} [h(x) \neq h'(x)] p_{\mathcal{ S}}(x) \mathrm{d} x$. The $\sup$ stands for the \emph{supremum}, which in this context finds the pair of classifiers $h,h'$ for which the difference in probability is largest and returns the value of that difference \cite{kifer2004detecting,ben2007analysis,ben2010theory}. 

\paragraph{} Given the error of the ideal joint hypothesis, $e^{*}_{\cal S,T} = \min_{h \in \mathcal{H}} \ [e_{\mathcal{ S}}(h) + e_{\mathcal{ T}}(h)]$, and the $\mathcal{ H}\Delta \mathcal{ H}$-divergence, a bound on the difference between the true target error, $e_{\cal T}$ of a trained source classifier, $\hat{h}_{\cal S} = \arg \min_{h} \hat{R}_{\cal S}(h)$, and that of the optimal target classifier, $h^{*}_{\cal T} = \arg \min_h R_{\cal T}(h)$, can be found. This bound has the following form (Theorem 3, \cite{ben2010theory}):
\begin{align}
	e_{\mathcal{ T}}(\hat{h}_{\mathcal{ S}}) - e_{\mathcal{ T}}(h_{\mathcal{ T}}^{*}) \ \leq e^{*}_{\cal S,T} + \frac{1}{2}d_{\mathcal{ H}\Delta \mathcal{ H}}(p_{\mathcal{ S}}, p_{\mathcal{ T}}) + \mathcal{ C}(\mathcal{H}) \, , \nonumber
\end{align}
which holds with probability $1-\delta$, for $\delta >0$. $\mathcal{ C}(\mathcal{H})$ describes the complexity of the type of classification functions $\mathcal{H}$ we are using, and comes up in standard generalization error bounds that incorporate classifier complexity \cite{vapnik1998statistical}. Overall, this bound states that, the larger $e^{*}_{\cal S,T}$ and $d_{\mathcal{ H}\Delta\mathcal{ H}}$ are for a given problem setting, the less a source classifier will generalize to the target domain.

\paragraph{} But the above bound describes the performance of a \emph{non-adaptive} classifier. Now, the challenge is to devise an adaptation strategy that leads to tighter generalization error bounds. This will not possible for the general case, but will be possible if the problem can be simplified through restrictions on the difference between domains. In the following, we discuss common simple data set shifts, for which adaptation strategies have been proposed with generalization error bounds.

\section{Common data shifts} \label{sec:shifts}
We are ultimately interested in minimizing the target risk $R_{\mathcal{ T}}$. So how does the source domain relate to this? One of the most straightforward ways to incorporate the source distribution in the target risk is as follows:
\begin{align}
	R_{\mathcal{ T}}(h) =& \sum_{y \in Y} \int_{\mathcal{ X}} \ell(h(x) \given y) \ p_{\mathcal{ T}}(x,y) \ \mathrm{d} x \nonumber \\
	=& \sum_{y \in Y} \int_{\mathcal{ X}} \ell(h(x) \given y) \ p_{\mathcal{ T}}(x,y) \ \frac{p_{\mathcal{ S}}(x,y)}{p_{\mathcal{ S}}(x,y)} \ \mathrm{d} x \nonumber \\
	=& \sum_{y \in Y} \int_{\mathcal{ X}} \ell(h(x) \given y) \ p_{\mathcal{ S}}(x,y) \ \frac{p_{\mathcal{ T}}(x,y)}{p_{\mathcal{ S}}(x,y)} \ \mathrm{d} x \, . \label{eq:imprisk1}
\end{align}
Which can be estimated using the sample average with:
\begin{align}
    \hat{R}_{\mathcal{ T}}(h) =& \frac{1}{n} \sum_{i=1}^{n} \ell(h(x_i), y_i) \ \frac{p_{\cal T}(x_i, y_i)}{p_{\cal S}(x_i, y_i)} \, . \nonumber
\end{align}
Note that the samples are drawn according to the source distribution, not the target distribution: $(x_i, y_i) \sim p_{\cal S}(x,y)$. $p_{\cal T}(x_i, y_i)$ describes the probability of those source samples under the target distribution. To estimate these probabilities, we would need labeled data from both domains, which is not available.

\paragraph{} Joint distributions can be decomposed in two ways: $p(x,y) = p(x \given y) p(y)$ and $p(x,y) = p(y \given x) p(x)$. If only one of the components differs between domains, then the resulting setting is a special case of data set shift. In $p(x,y) = p(x \given y) p(y)$, the class-conditional distributions are weighted by the prior distribution, i.e. the relative proportion of each class. If the conditional distributions remain constant and only the prior distributions differ, then it is said to be a case of "prior shift". In $p(x,y) = p(y \given x) p(x)$, there are two special cases. In the first, the posterior distributions are equivalent but the data distributions differ between domains. This is known as "covariate shift". If the data distributions remain constant, but the posteriors change, then it is a case of "concept shift". For prior and covariate shift, it is not necessary to obtain labeled data in both domains. The following subsections discusses these three simple types of shifts in more detail.

\subsection{Prior shift}
For prior shift, the prior probabilities of the classes are different, $p_{\mathcal{ S}}(y) \neq p_{\mathcal{ T}}(y)$, but the conditional distributions are equivalent, $p_{\mathcal{ S}}(x|y) = p_{\mathcal{ T}}(x|y)$. This can occur in for example fault detection settings, where a new maintenance policy might cause less faults \cite{japkowicz1995novelty}, or in the detection of oil spills before versus after an incident \cite{kubat1998machine}. 

\begin{figure}[htb]
\includegraphics[width=.31\textwidth]{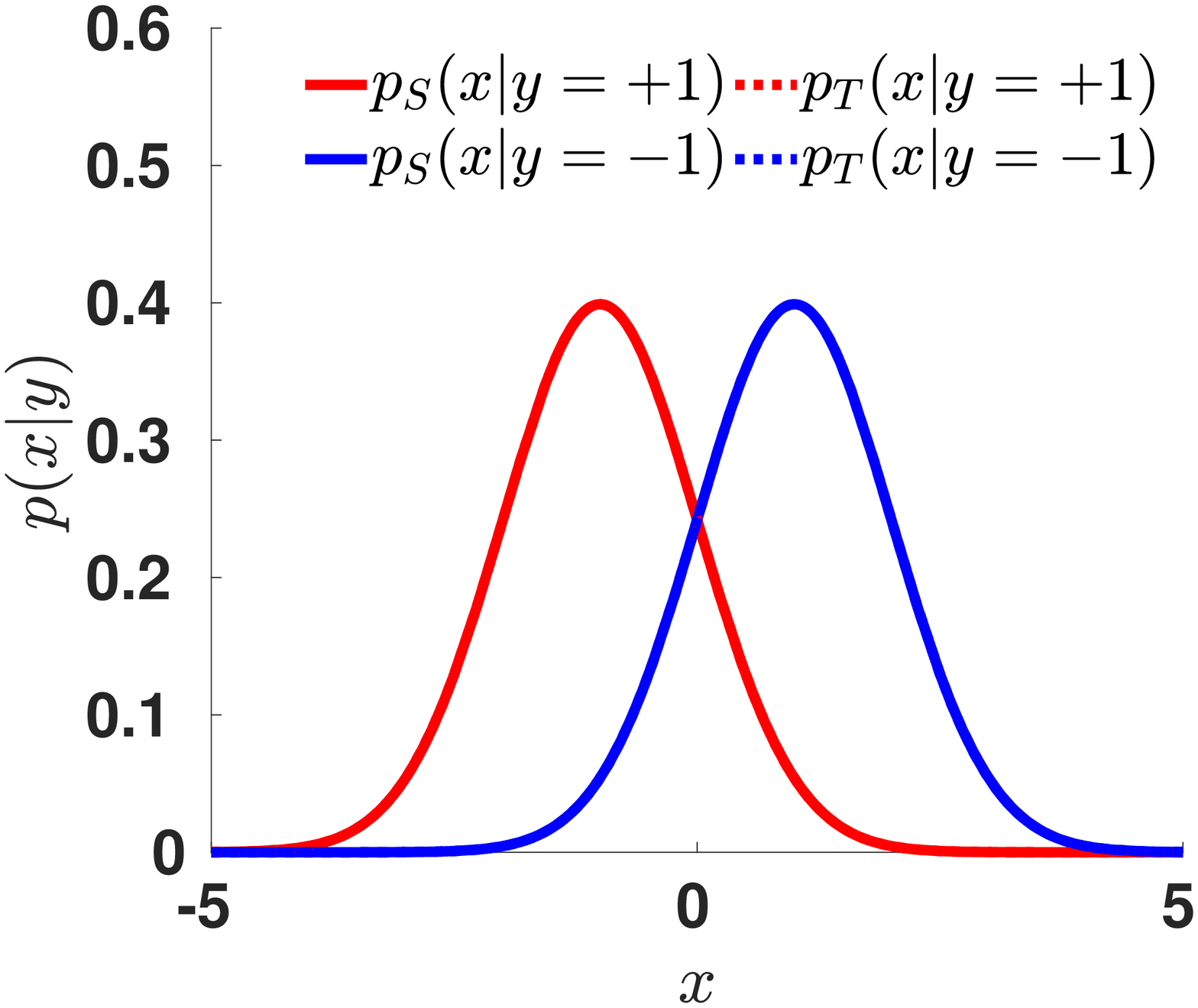}\
\includegraphics[width=.31\textwidth]{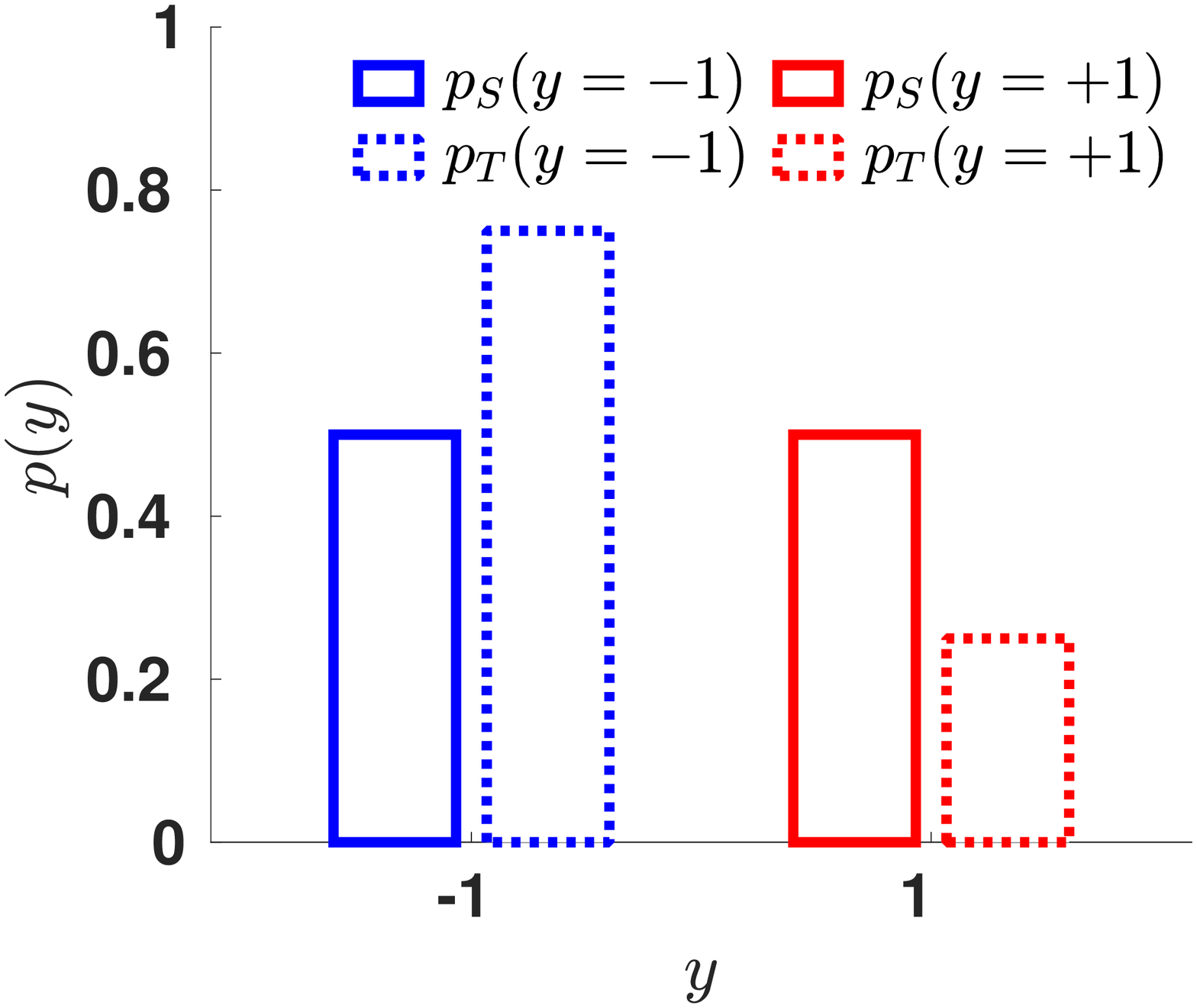}\
\includegraphics[width=.31\textwidth]{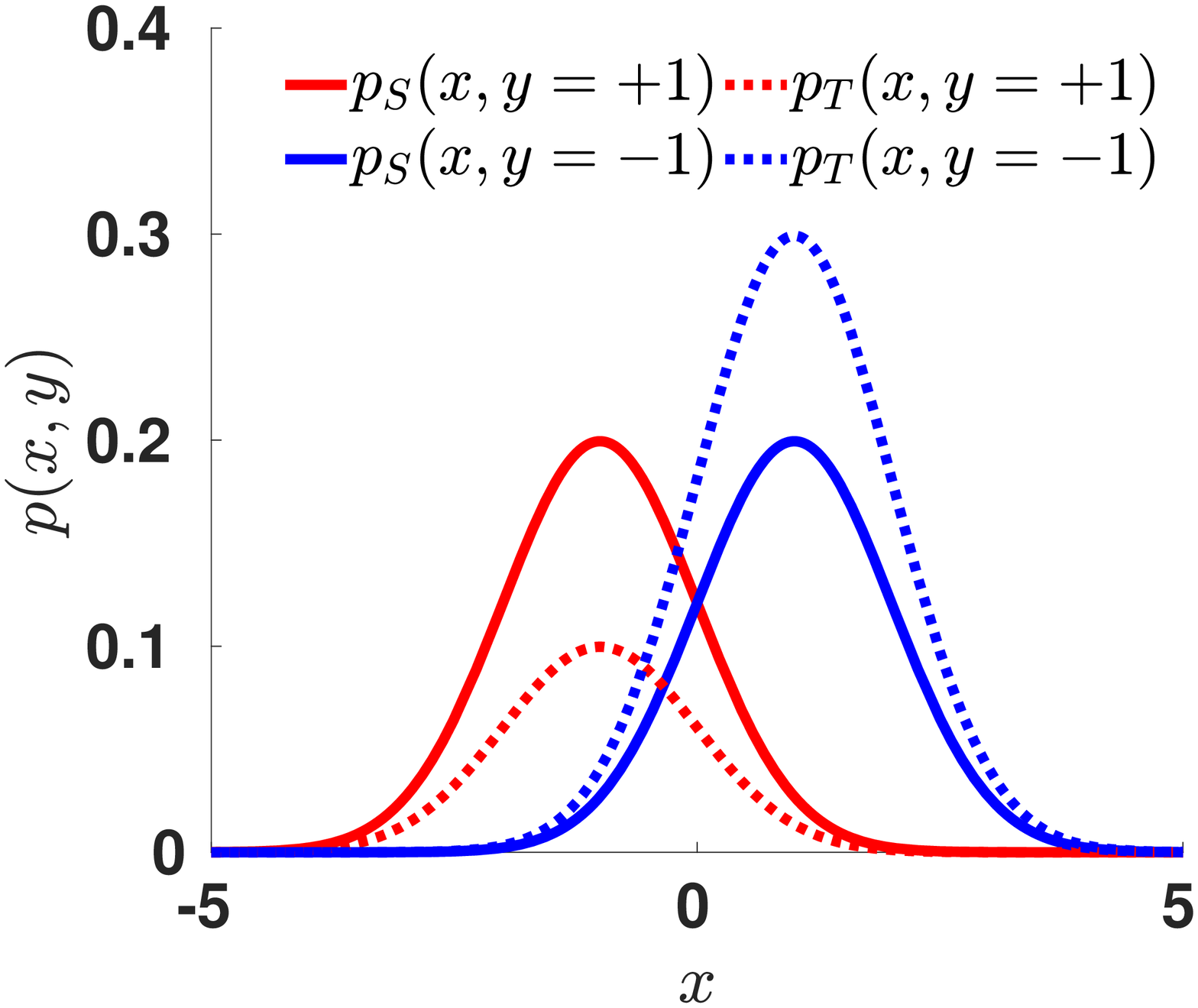}
\caption{An example of prior shift. (Left) The conditional distributions in each domain are different. (Middle) The prior distribution changes from $1/2$ for both classes in the source domain to $3/4$ and $1/4$ in the target domain. (Right) The joint distribution for the source domain is balanced, while the negative class outweighs the positive class in the target domain.}
\label{fig:priorshift}
\end{figure}

\paragraph{} Figure \ref{fig:priorshift} illustrates an example of prior shift. The prior probabilities for the positive and negative class are both $1/2$ in the source domain, but $2/3$ versus $1/3$ in the target domain. The data and posterior distributions in each domain remain equivalent, which results in a change in the conditional distributions. The positive class outweighs the negative class in probability. 

\paragraph{} The knowledge that the conditionals are equivalent can be exploited by canceling them out of the ratio of joint probability distributions \cite{saerens2002adjusting}:
\begin{align}
	R_{\mathcal{ T}}(h) =& \sum_{y \in Y} \int_{\mathcal{ X}} \ell(h(x), y) \frac{\cancel{p_{\mathcal{ T}}(x \given y)} \ p_{\mathcal{ T}}(y)}{\cancel{p_{\mathcal{ S}}(x \given y)} \ p_{\mathcal{ S}}(y)} p_{\mathcal{ S}}(x,y) \ \mathrm{d} x \, , \label{eq:prior}
\end{align}
where the ratio $p_{\mathcal{ T}}(y) / p_{\mathcal{ S}}(y)$ represent the change in class proportions. Using samples drawn from the source distribution and a function that re-weights each class, we can estimate the target risk using the following estimator:
\begin{align}
    \hat{R}_{\mathcal{ T}}(h) =& \frac{1}{n} \sum_{i=1}^{n} \ell(h(x_i), y_i) \ w(y_i) \, . \nonumber
\end{align}
Using this approach, no unlabeled target samples are necessary, only target labels.

\paragraph{} Class-based weighting has been extensively studied from an alternative perspective: when it is more difficult to collect data from one class than the other \cite{he2009learning}. For example, in a few countries, women above a certain age are given the opportunity to be tested for breast cancer \cite{scaf1997screening}. The vast majority that responds does not show signs of cancerous tissue and only a small minority is tested positive. On this data set, the classifier could always predict 'healthy' and achieve a low error. But then it misses exactly the cases of interest. To avoid this unwanted behaviour, samples from the rare class could be re-weighted to be more important \cite{elkan2001foundations,chawla2002smote,estabrooks2004multiple}.

\subsection{Covariate shift}
Covariate shift is one of the most studied forms of data set shift. It occurs most often when there is a form of sample selection bias \cite{heckman1977sample,lee1982some,helton2006survey}. Selection bias is defined as the altered probability of being sampled \cite{heckman1990varieties,cochran1973controlling}. For example, suppose you would visit a city where most people live in the center and the habitation density decreases as a function of the distance from the center. You are interested in whether people think that the city is overpopulated. If you would sample on the main square you will mostly encounter people who live in the center, and you would likely get a lot of 'yes' answers. Inhabitants who live further away, who would say 'no', are under-represented in the data. The results of your survey will likely differ from if you would sample door-to-door. 

From a domain adaptation perspective, the biased sampling corresponds to the source domain and the target domain to the unbiased sample. Re-weighting individual samples would correspond to correcting the over-representation of people living in the center and under-representation of people living further away. 

\paragraph{} Similar to sample selection bias, another cause for covariate shift is missing data \cite{rubin1976inference,little2014statistical}. In practice, data can be missing as measurement devices fail or because of subject dropout. When there is a consistent mechanism behind how the data went missing, referred to as missing-not-at-random (MNAR), the missingness constitutes an additional variable. The collected results are dependent on the probability of 'not-missing', and adaptation consists of correcting for under- or over-observed samples.

\begin{figure}[htb]
\includegraphics[width=.31\textwidth]{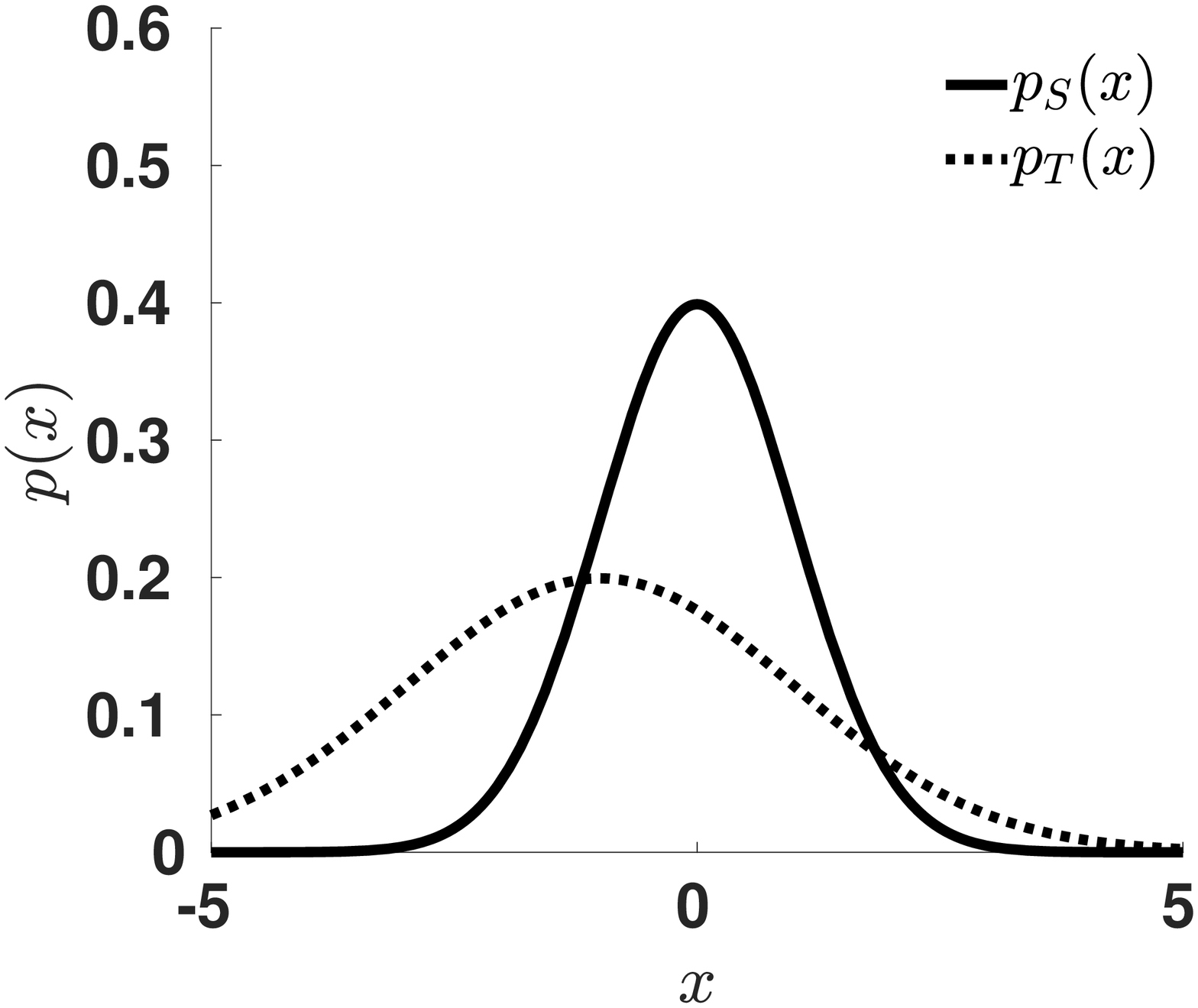}\
\includegraphics[width=.31\textwidth]{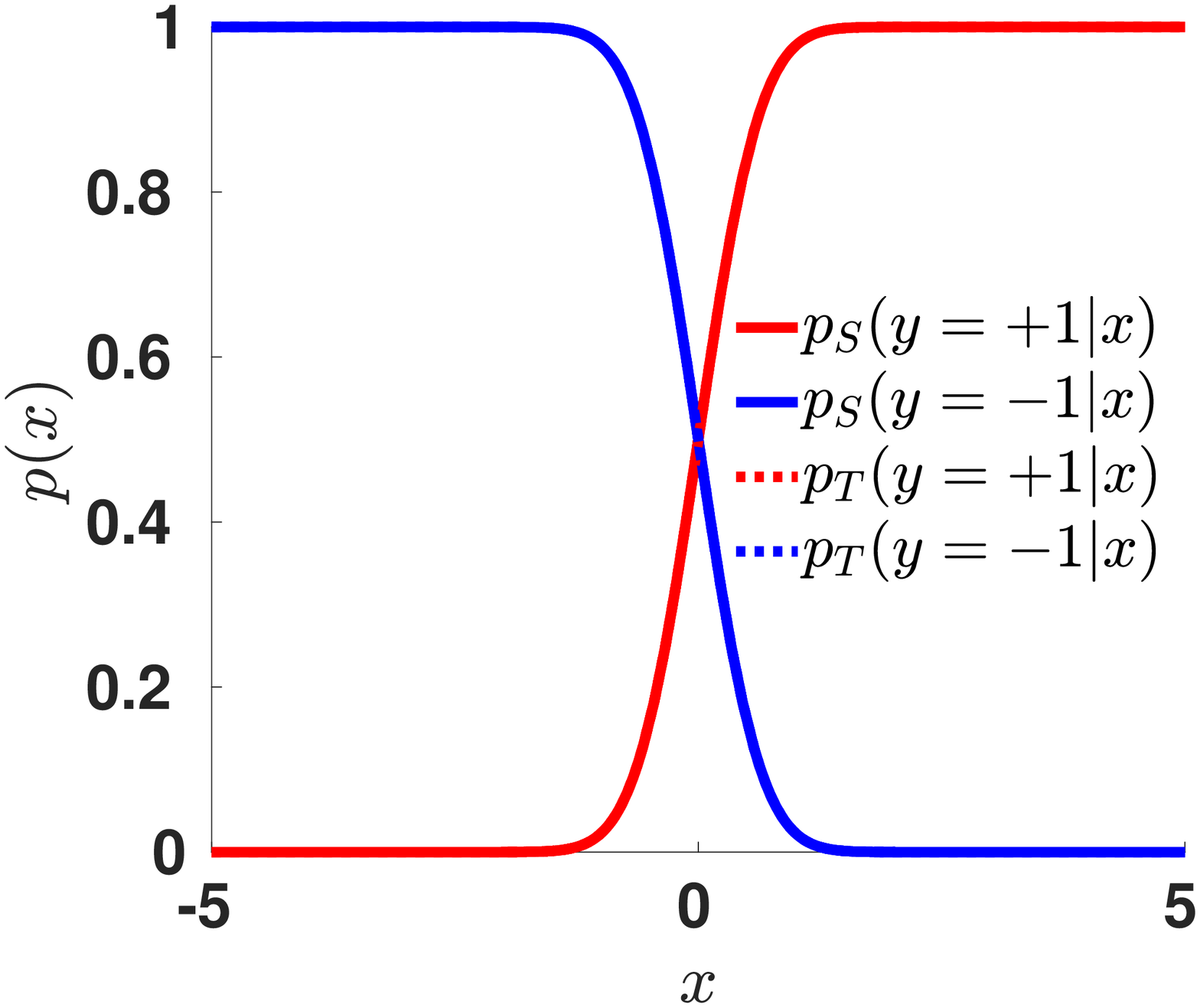}\
\includegraphics[width=.31\textwidth]{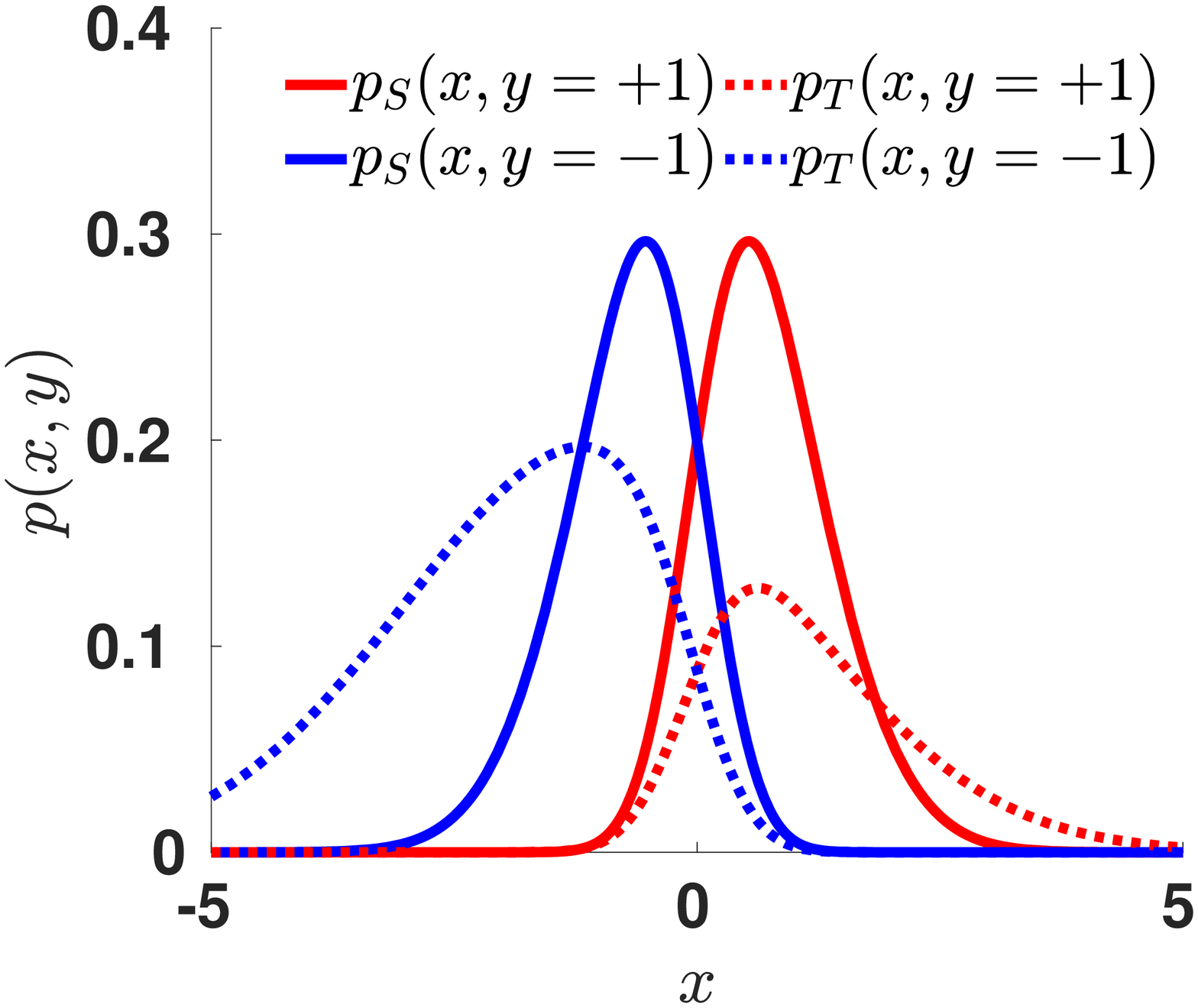}
\caption{An example of covariate shift. (Left) The target data distribution has shifted away from the source data distribution. (Middle) The posterior distributions are equal. (Right) The resulting joint distributions are shifted as well.}
\label{fig:covariateshift}
\end{figure}

\paragraph{} Figure \ref{fig:covariateshift} presents an example of a case of covariate shift. On the left are shown the data distributions in each domain. The source distribution is centered on zero, while the target is centered on $-1$. The posterior distributions are the same (middle), but the joint distributions differ.

\paragraph{} The knowledge that the posterior distributions are equivalent can be exploited by canceling them in the ratio of joint distributions in  (\ref{eq:imprisk1}):
\begin{align}
	R(h) =& \sum_{y \in Y} \int_{\mathcal{ X}} \ell(h(x), y) \frac{\cancel{p_{\mathcal{ T}}(y \given x)} \ p_{\mathcal{ T}}(x)}{\cancel{p_{\mathcal{ S}}(y \given x)} \ p_{\mathcal{ S}}(x)} p_{\mathcal{ S}}(x,y) \ \mathrm{d} x \, , \label{eq:covariate}
\end{align}
where the ratio $p_{\mathcal{ T}}(x) / p_{\mathcal{ S}}(x)$ indicates how the probability of a source sample should be corrected to reflect the probability under the target distribution.

\subsection{Concept shift}
In the case of concept shift, the data distributions remain constant while the posteriors change. For instance, consider a medical setting where the aim is to make a prognosis for a patient based on their age, severity of their flu, general health and their socio-economic status. In \cite{alaiz2008assessing}, the classes are originally defined as "remission" and "complications". But, at test time, other aspects are counted as a form of "complication" and are so labeled. What constitutes the positive and negative class, and by extension the posterior distributions, has changed.

\begin{figure}[htb]
\includegraphics[width=.31\textwidth]{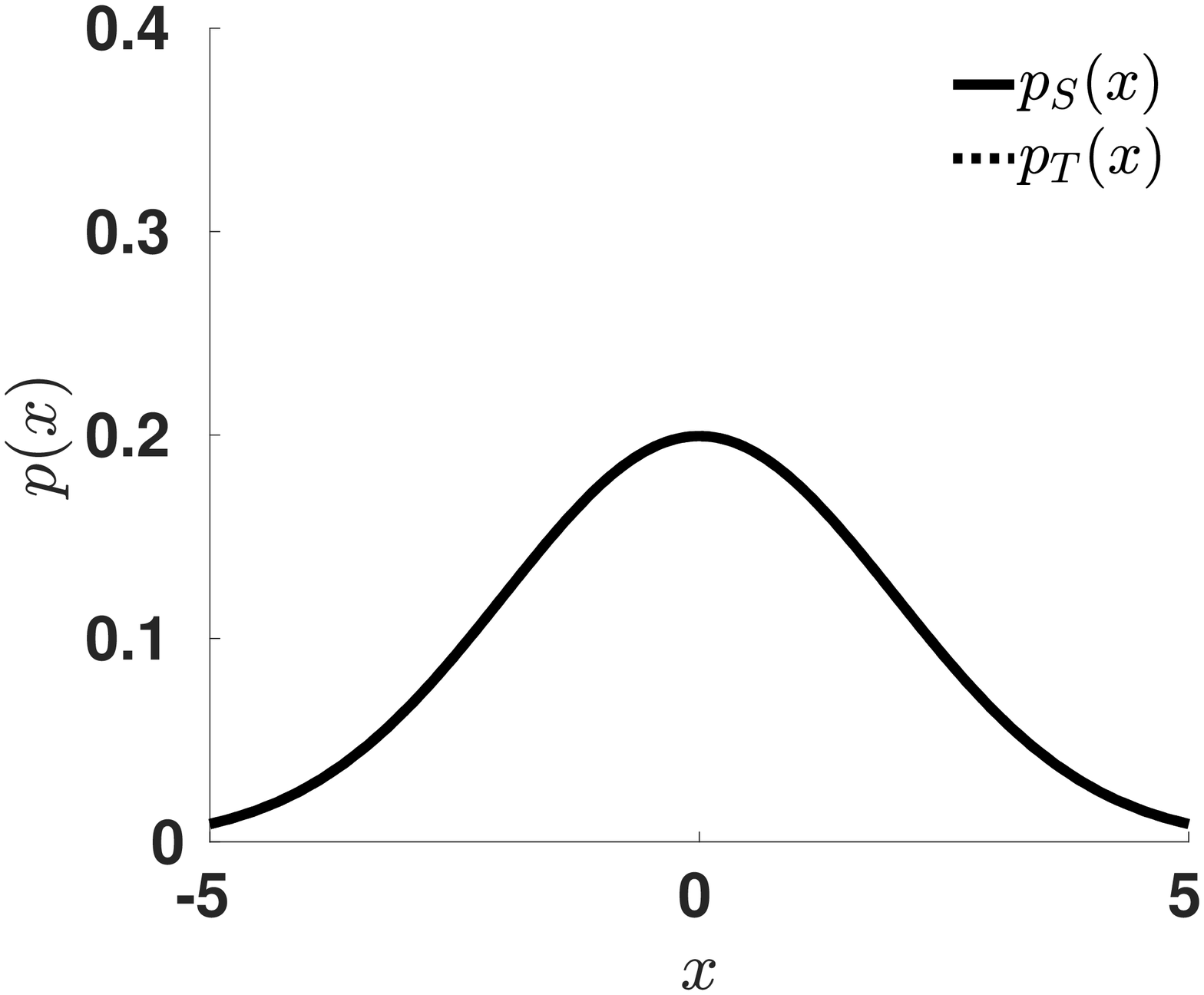}\
\includegraphics[width=.31\textwidth]{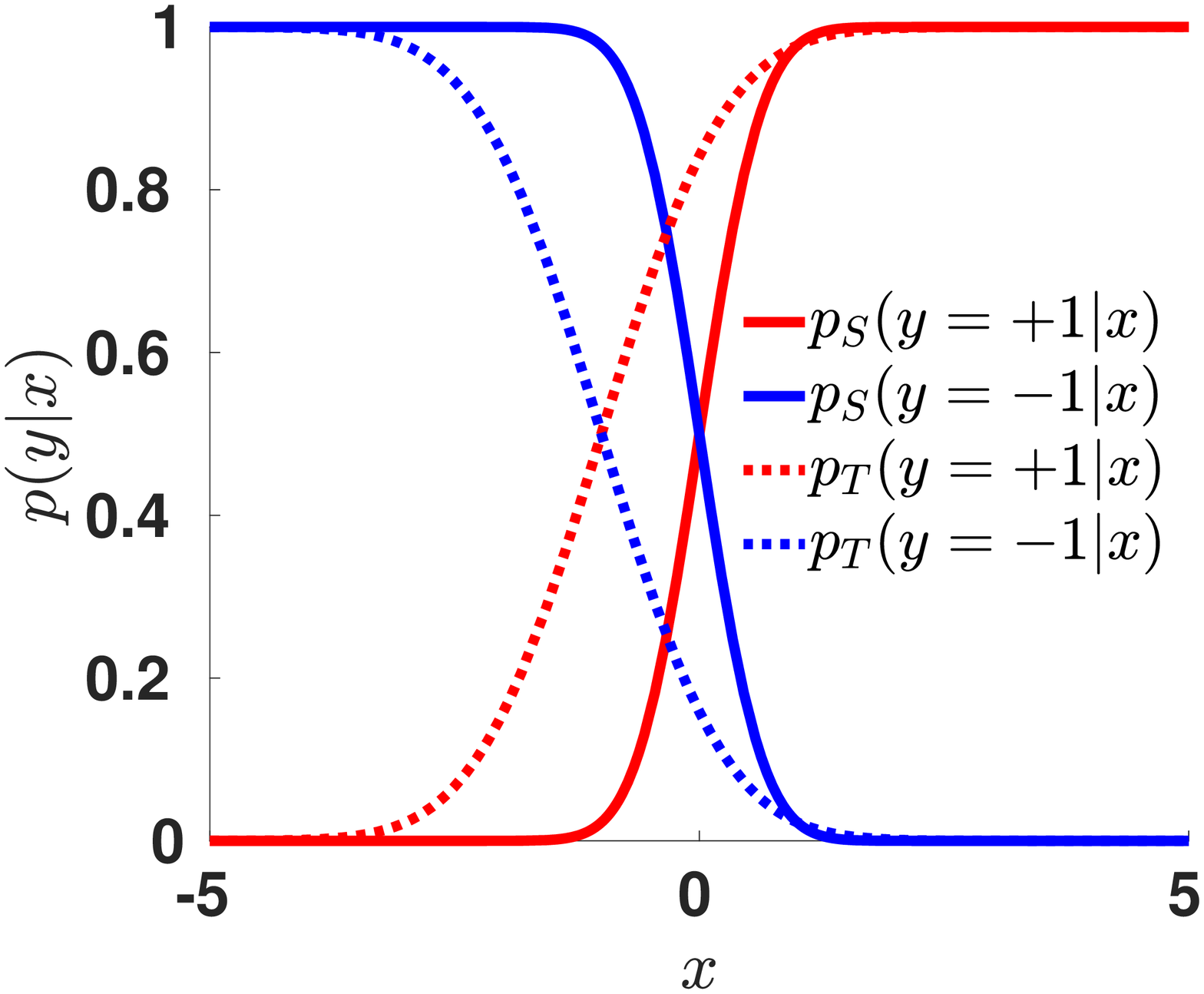}\
\includegraphics[width=.31\textwidth]{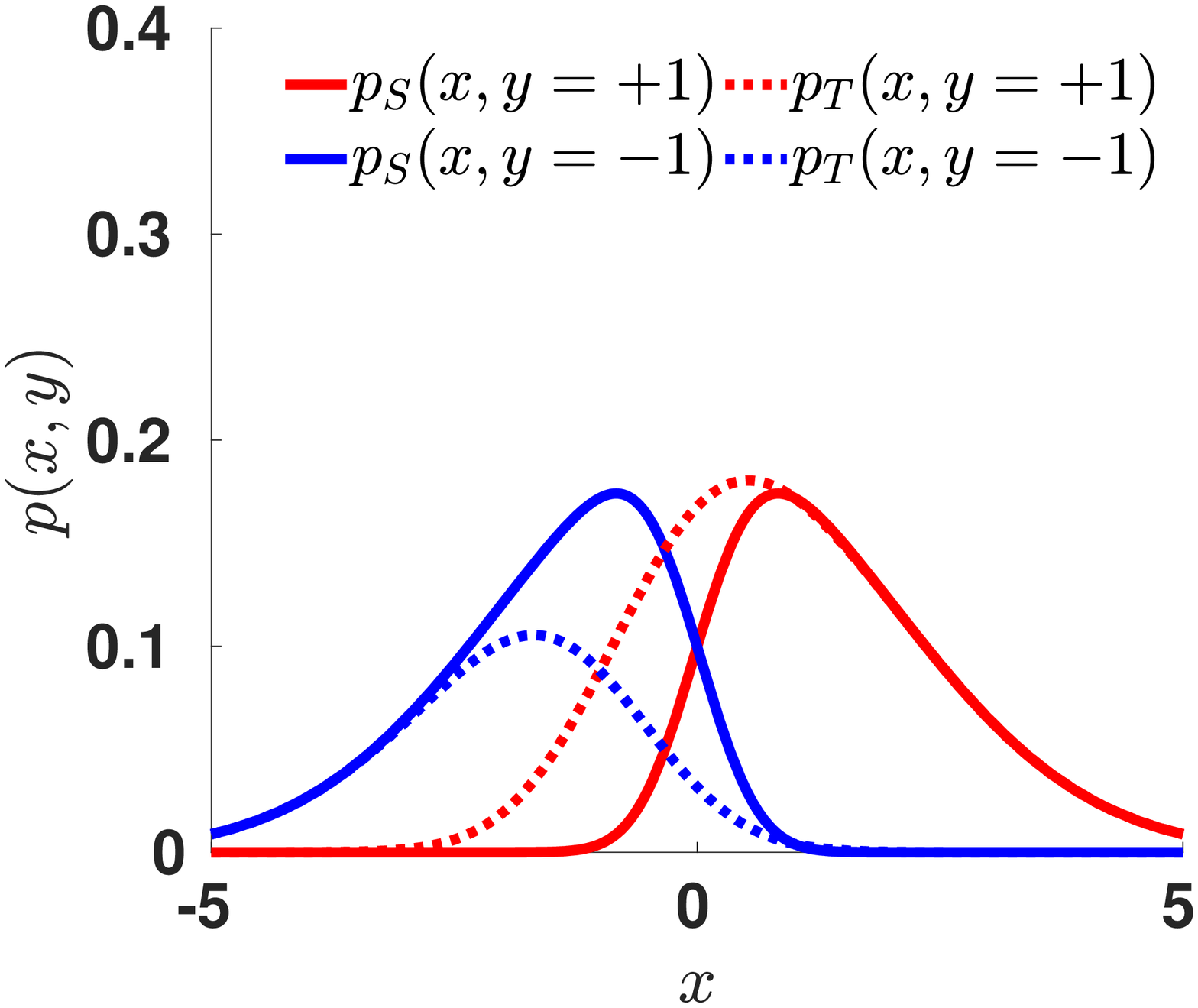}
\caption{An example of concept shift. (Left) The data distributions remain constant. (Middle) The target posterior distributions are shifted to the left of the source posteriors. (Right) The resulting joint distributions are shifted as well.}
\label{fig:conceptshift}
\end{figure}

\paragraph{} Figure \ref{fig:conceptshift} shows an example of setting with concept shift. In this case, the posterior distribution in the target domain has shifted away from the source distribution, towards the negative part of feature space. In the resulting joint distribution, the decision boundary has shifted to the left as well.

\paragraph{} The knowledge that the data distributions remain equal can be exploited through:
\begin{align}
	R_{\mathcal{ T}}(h) =& \sum_{y \in Y} \int_{\mathcal{ X}} \ell(h(x), y) \frac{p_{\mathcal{ T}}(y \given x) \ \cancel{p_{\mathcal{ T}}(x)}}{p_{\mathcal{ S}}(y \given x) \ \cancel{p_{\mathcal{ S}}(x)}} p_{\mathcal{ S}}(x,y) \ \mathrm{d} x \nonumber \, .
\end{align}
But adaptation in this setting is still impossible without labeled target data. To estimate conditional distributions, one requires simultaneous observations of both variables. 

\paragraph{} Concept shift is related to \emph{data drift}, where classifiers are deployed in non-stationary environments \cite{widmer1996learning}. For smoothly varying non-stationarities, such as time-series, there is again additional information that can be exploited: the shifts are ordered and are relatively small between neighboring time steps. If the classifier receives feedback after each time step, the drift can be modeled and the next timestep's shift can be predicted \cite{dietterich2002machine,gama2006learning,ditzler2011hellinger,chen2015sequential}.

\subsection{General domain shifts} \label{sec:gen_dom}
In general cases of domain shift, one or more of the above shifts will have occurred. As can be imagined, this is the most difficult setting and learning will often not be possible at all \cite{ben2010impossibility,ben2012hardness}. In order to generalize well, the domains have to be related in some other exploitable way. Examples of exploitable relationships include: the existence of a single good predictor for both domains \cite{ben2007analysis,ben2010impossibility,ben2010theory,blitzer2011domain}, constrained worst-case labellings \cite{wen2014robust,liu2014robust}, low-data-divergence \cite{ben2007analysis,ben2010impossibility,ben2010theory}, the existence of a domain manifold \cite{gopalan2011domain,baktashmotlagh2014domain,patel2015visual}, conditional independence of class and target given source data \cite{kouw2016feature} and unconfoundedness \cite{imbens2015causal}. Some of these domain relationships are discussed in the following Section.

\section{Approaches} \label{sec:approaches}
There are many approaches to transfer learning and domain adaptation. Some of the more prominent ones are discussed here.

\subsection{Importance-weighting} \label{sec:iw}
Weighting samples based on their importance to the target domain is mostly used in covariate shift. Although the step in Equation \ref{eq:covariate} seems straightforward, one still needs to show that an importance-weighted classifier will learn, i.e. improve with more samples. For that, a generalization error bound is needed \cite{cortes2010learning}. The difference between the true target error of a classifier, $e_{\cal T}(h)$, and the empirical weighted source error, $\hat{e}_{\cal W}(h)$, is (c.f. Theorem 3, \cite{cortes2010learning}):
\begin{align}
e_{\cal T}(& h) - \hat{e}_{\cal W}(h) \leq \ 2^{5/4} \ \sqrt{\mathrm{D}_{2}( p_{\cal T} \| p_{\cal S})} \ \sqrt[3/8]{\frac{c}{n} \log\frac{2 n e}{c} + \frac{1}{n} \log \frac{4}{\delta}} \label{eq:geb_iw} \, ,
\end{align}
which holds with probability at least $1- \delta$, for $\delta > 0$. $\mathrm{D}_{2R}( p_{\cal T} \| p_{\cal S})$ is the 2-order R{\'e}nyi divergence \cite{van2014renyi}, an information-theoretic domain discrepancy measure, and $c$ refers to the pseudo-dimension of $\mathcal{H}$, a measure of describing the complexity of the hypothesis space \cite{vidyasagar2002theory}. $c$ and $\mathrm{D}_{2R}( p_{\cal T} \| p_{\cal S})$ are required to be finite and the weights cannot be equal to $0$. $\mathrm{D}_{2R}( p_{\cal T} \| p_{\cal S})$ will diverge to infinity when the domains are too far apart, in which case the above generalization error bound does not hold.

We can now see that the difference between the adaptive classifier obtained using importance-weighting depends on the complexity of the classifier, the sample size and the divergence between the domains. In other words: for a fixed choice of hypothesis space (e.g. linear classifiers), as the divergence between the domains increases, the sample size needs to increase at a certain rate as well, in order to maintain the same probability of being approximately correct.

\subsubsection{Weight estimation} 
Given that importance-weighting is a valid strategy for domain adaptation, the question is how to estimate importance weights appropriately. Depending on the problem setting, some methods estimate the numerator and denominator of the ratio of probabilities separately, and others estimate the ratio directly.

\paragraph{} Each probability distribution in the ratio can be estimated using a Gaussian distribution \cite{shimodaira2000improving}. But this tends to have a negative effect on the variance of the importance weights \cite{cortes2008sample,cortes2010learning,cortes2014domain}. For example, suppose the source distribution is a univariate Gaussian distribution with mean $0$ and variance $1$, and the target distribution is a univariate Gaussian with mean $0$ and variance $\sigma^2_{\mathcal{ T}}$. Then, the weights consist of $p_{\mathcal{ T}}(x) / p_{\mathcal{ S}}(x) = \ \mathcal{N}(x \given 0, \sigma^{2}_{\mathcal{ T}}) \ / \ \mathcal{N}(x \given 0, 1) =  \ \sigma^{-1}_{\mathcal{ T}} \exp (x^{2} (-1 + \sigma^{2}_{\mathcal{ T}} ) / (2\sigma^{2}_{\mathcal{ T}}) )$. If the target variance is larger than $2$, then the variance of the weights, $\mathbb{E}_{\mathcal{ S}} [ ( w(x) - \mathbb{E}_{\mathcal{ S}}[w(x)])^2]$, diverges to infinity. Large weight variance means that it is highly probable that one sample will receive a very large weight, while the rest will receive nearly-zero weights. Consequently, at training time, the classifier will focus on this one important sample and will effectively ignore the others. The resulting classifier is often pathological and will not generalize well. 

\paragraph{} A non-parametric alternative is to use kernel density estimation (KDE) \cite{bickel2009discriminative,yu2012analysis,baktashmotlagh2014domain}. In KDE, a distribution is estimated by placing a kernel function on top of each sample. This kernel function is $1$ in its center and drops off exponentially based on the distance away from the center. The probability density of any point in the space is the average kernel distance to all known points. KDE can be used to estimate weights by first estimating the density of a source sample with respect to all target samples. Secondly, the density of the source samples is estimated and lastly, the target density is divided by the source density. 

The advantage of KDE is that it is possible to control the variance properties of the resulting ratio through the design of the kernel. It contains what is known as a \emph{kernel bandwidth} parameter. This bandwidth parameter roughly corresponds to how wide the kernel is. It is a hyperparameter, and although it gives some control, it is not clear how it should be set.

\paragraph{} Instead of estimating the data distribution in each domain separately, one could estimate the ratio directly. Methods that directly estimate importance weights are usually based on minimizing some type of discrepancy between the weighted source and the target distributions: $\mathrm{D}\left[w, p_{\mathcal{ S}}, p_{\mathcal{ T}} \right]$ \cite{sugiyama2012density}. However, just minimizing this objective with respect to $w$ might cause highly varying or unusually scaled values \cite{tsuboi2009direct}. This unwanted behaviour can be combated through incorporating a property of the reweighed source distribution:
\begin{align} 
	1 =& \ \int_{\mathcal{ X}} p_{\mathcal{ T}}(x) \mathrm{d} x \nonumber \\
	=& \ \int_{\mathcal{ X}} w(x) p_{\mathcal{ S}}(x) \mathrm{d} x \nonumber \\
	\approx& \ \frac{1}{n} \sum_{i=1}^{n} w(x_i) \, , \label{eq:avgto1}
\end{align}
for samples drawn from the source distribution, $x_i \sim p_{\mathcal{ S}}$. Restraining the weight average to be close to $1$, disfavors large values for weights. The approximate equality can be enforced by constraining the absolute deviation of the weight average to $1$ to be less than some small value: $| \  n^{-1} \sum_{i}^{n} w(x_i) - 1 \ | \leq \epsilon$.  

Incorporating the average weight constraint, along with the constraint that the weights should all be non-negative, direct importance weight estimation can be formulated as the following optimization problem:
\begin{align}
	\underset{w \in W}{\text{minimize}} \quad& \mathrm{D}\left[w,  p_{\mathcal{ S}}, p_{\mathcal{ T}} \right] \nonumber \\
	\text{s.t.} \quad& w_i \geq 0 \nonumber \\
	\quad& | \  \frac{1}{n} \sum_{i}^{n} w_i - 1 \ | \leq \epsilon \, .	\label{eq:nonparam_iw}
\end{align}
Note that $w$ is now a variable and not a function of the sample $w(x)$. Depending on the choice of discrepancy measure, this optimization problem could be linear, quadratic or be even further constrained.

\paragraph{} A popular measure of domain discrepancy is the Maximum Mean Discrepancy, which is based on the two-sample problem from statistics \cite{fortet1953convergence,borgwardt2006integrating,gretton2007kernel}. It measures the distance between two means after subjecting the samples to the continuous function that pulls them maximally apart. In practice, kernel functions are used, which, under certain conditions, are able to approximate any continuous function arbitrary well \cite{aronszajn1950theory,scholkopf2002learning,shawe2004kernel,berlinet2011reproducing}. The empirical discrepancy measure, including the reweighed source samples, can be expressed as \cite{gretton2009covariate}:
\begin{align}
\hat{\mathrm{D}}^2_{\text{MMD}}[w,X,Z] =& \ \frac{1}{n^{2}} \sum_{i,i'}^{n} w_i\kappa(x_i,x_{i'}) w_{i'} - \frac{2}{mn} \sum_{i}^{n} \sum_{j}^{m} w_i \kappa(x_i,z_j) + C \, , \label{eq:kmm}
\end{align}
where $\kappa$ are the kernel functions and $C$ are constants not relevant to the optimization problem. Minimizing the empirical MMD with respect to the importance weights, is called Kernel Mean Matching (KMM) \cite{huang2007correcting,gretton2009covariate}. Depending on if the weights are upper bounded, convergence rates for KMM can be computed \cite{gretton2007kernel,gretton2009covariate,yu2012analysis}

\paragraph{} Another common measure of distribution discrepancies is the Kullback-Leibler Divergence \cite{kullback1951information,cover2012elements,mackay2003information}, leading to the Kullback-Leibler Importance Estimation Procedure (KLIEP) \cite{sugiyama2005model,sugiyama2007covariate,sugiyama2008direct}. The KL-divergence between the true target distribution and the importance-weighted source distribution can be simplified as:
\begin{align}
	\mathrm{D}_{\text{KL}} \left[ w,  p_{\mathcal{ S}}, p_{\mathcal{ T}} \right] =& \int_{\mathcal{ X}} p_{\mathcal{ T}}(x) \log \frac{p_{\mathcal{ T}}(x)}{p_{\mathcal{ S}}(x) w(x)} \mathrm{d} x \nonumber \\
	=& \int_{\mathcal{ X}} p_{\mathcal{ T}}(x) \log \frac{p_{\mathcal{ T}}(x)}{p_{\mathcal{ S}}(x)} \mathrm{d} x  - \int_{\mathcal{ X}} p_{\mathcal{ T}}(x) \log w(x) \mathrm{d} x \, . \label{kliep1}
\end{align}
Since the first term in the right-hand side of (\ref{kliep1}) is independent of $w$, only the second term is used as in the optimization objective function. This second term is the expected value of the logarithmic weights with respect to the target distribution, which can be approximated with unlabeled target samples: $\mathbb{E}_{\mathcal{ T}}[ \ \log w(x) ] \approx m^{-1} \sum_{j}^{m} \log w(z_j)$. The weights are formulated as a functional model consisting of an inner product of weights $\alpha$ and basis functions $\phi$, i.e. $w(x) = \alpha^{\top} \phi(x)$ \cite{sugiyama2005input}. This allows them to apply the importance-weight function to both the test samples in the KLIEP objective from (\ref{kliep1}) and to the training samples for the constraint in (\ref{eq:avgto1}). 

\paragraph{} One could also minimize the squared error between the estimated weights and the actual ratio of distributions \cite{kanamori2009least,kanamori2009efficient}: 
\begin{align}
\mathrm{D}_{\text{LS}}[w,p_{\mathcal{ S}}, p_{\mathcal{ T}}] =& \frac{1}{2} \int_{\mathcal{ X}} \left(w(x) - \frac{p_{\mathcal{ T}}(x)}{p_{\mathcal{ S}}(x)} \right)^2 p_{\mathcal{ S}}(x) \mathrm{d} x \nonumber \\
=& \frac{1}{2} \int_{\mathcal{ X}} w(x)^2 p_{\mathcal{ S}}(x) \mathrm{d} x - \int_{\mathcal{ X}} w(x) p_{\mathcal{ T}}(x) \mathrm{d} x + C \, , \label{LSIF}
\end{align} 
where $C$ is again a term containing constants not relevant to the optimization problem. As this squared error is used as an optimization objective function, the constant term drops out. We are then left with the expected value of the squared weights with respect to the source distribution, and the expected value of the weights with respect to the target distribution. Expanding the weight model, $w(x) = \alpha^{\top} \phi(x)$, gives $1/2 \ \alpha^{\top} \mathbb{E}_{\mathcal{ S}} [\phi(x) \phi(x)^{\top} ] \alpha -   \mathbb{E}_{\mathcal{ T}} [\phi(x) ]$. Replacing the expected values with sample averages allows for plugging in this objective into the nonparametric weight estimator in (\ref{eq:nonparam_iw}). This technique is called Least-Squares Importance Fitting (LSIF).


\paragraph{} Lastly, directly estimating importance weights can be done through tessellating the feature space into Voronoi cells \cite{silverman1986density,okabe1992spatial,loog2012nearest,kremer2015nearest}. Each cell is a polygon of variable size and denotes an area of equal probability \cite{miller2003new,learned2004hyperspacings}. The cells approximate a probability distribution function in the same way that a multi-dimensional histogram does: with more Voronoi cells, one obtains a more precise description of the change in probability between neighbouring samples. To estimate importance weights, first, one forms the Voronoi cell $V_i$ of each source sample $x_i$. The cell consists of the part of feature space that lies closest to $x_i$ and can be obtained through a 1-nearest-neighbour procedure. The ratio of target over source is then approximated by counting the number of target samples $z_j$ that lie within the cell: $w(x_i) = |V_i \cap \{z_j\}_{j=1}^{m} |$, where $\cup$ denotes the intersection between the Voronoi cell and the set of target samples and $| \cdot |$ denotes the cardinality of this set. 

It does not require hyperparameter optimization, but there is still the option to perform Laplace smoothing, where a 1 is added to each cell \cite{field1988laplacian}. This ensures that no source samples are given a weight of $0$ and are thus completely discarded.

\subsection{Subspace mappings} \label{sec:submappings}
Domains may lie in different subspaces \cite{torralba2011unbiased,fernando2013unsupervised}, in which case there would exist a mapping from one domain to the other \cite{gopalan2015domain,patel2015visual}. For example, the mapping may correspond to a rotation, an affine transformation, or a more complicated nonlinear transformation \cite{fernando2013unsupervised,pan2011domain}. In problems in computer vision and natural language processing, the data can be very high-dimensional and the domains might look completely different from each other. For example, photos of animals in zoos versus pictures of animals online. Unfortunately, using a too flexible transformation to capture this mapping, can lead to overfitting. This means the method will work well for the given target samples but fail for new target samples. Note that any structural relationships between domains, such as equal posterior distributions will no longer be valid if the domains are mapped separately.

\paragraph{} The simplest technique for finding a subspace mapping is to take the principal components in each domain, $C_{\mathcal{ S}}$ and $C_{\mathcal{ T}}$, and rotate the source components to align with the target components: $C_{\mathcal{ S}} W$  \cite{fernando2013unsupervised}. $W$ is the linear transformation matrix, which consists of the transposed source components times the target components $W = C_{\cal S}^{\top} C_{\cal T}$. However, each domain contains noise components and these should not affect the matching procedure. Subspace Alignment (SA) can be used to find an optimal subspace dimensionality \cite{fernando2013unsupervised}. This technique is attractive because its flexibility is limited, making it quite robust to unusual problem settings. It is computationally not expensive, easily implemented and intuitive to explain. It has been extended a number of times: there is a landmark-based kernelized alignment \cite{aljundi2015landmarks}, a subspace distribution alignment technique \cite{sun2015subspace} and a semi-supervised variant \cite{yao2015semi}. 

\paragraph{} Instead of aligning the domains based on directions of variance, one could model the structure of the data with graph-based methods \cite{das2018unsupervised,das2018graph}. Data is first summarized using a subset called the exemplars, obtained through clustering. From the set of exemplars, two sets of hyper-graphs are constructed. These two hyper-graphs are matched along their first, second and third orders \cite{das2018unsupervised,das2018graph}. First-order matching finds a linear transformation matrix that minimizes the difference between the mapped source and the target hyper-graphs, similar to Subspace Alignment. Higher-order moments consider other forms of geometric and structural information, beyond pairwise distances between exemplars \cite{das2018sample,das2018unsupervised}. These can be found using tensor-based algorithms \cite{das2018unsupervised,duchenne2011tensor}.

\subsubsection{Metrics} A number of methods aim to find transformations that aid specific subsequent classifiers \cite{saenko2010adapting,kulis2011you,geng2011daml}. For example, Information-Theoretic Metric Learning (ITML) \cite{saenko2010adapting}. In ITML, a Mahalanobis metric is learned for use in a later nearest-neighbour classifier \cite{davis2007information,saenko2010adapting}. Metrics describe ways of computing distances between points in vector spaces. The standard Euclidean metric, $d^2_{E}(x,z) = (x-z)(x-z)^{\top}$, consists of the inner product of the difference between two vectors. The Mahalanobis metric, $d^2_{W}(x,z) = (x-z)^{\top}W(x-z)$, weights the inner product by the curvature of the space $W$. In fact, first transforming the space and then measuring distances with the Euclidean metric is equivalent to measuring distances with the Mahalanobis metric according to the square root of the transformation matrix $W$: $(W^{1/2}x - W^{1/2}z)^{\top}(W^{1/2}x - W^{1/2}z) = (x-z)^{\top} W (x-z)$. Hence, first transforming the space and then training a nearest-neighbour classifier is equivalent to training a Mahalanobis nearest-neighbour.

In order to avoid that the Mahalanobis metric pushes samples from different classes closer together for the sake of mapping source to target, it is necessary to include some constraints \cite{bar2005learning}. If a small number of target labels is available, then these could be used as correspondence constraints. They would consist of thresholding the pairwise distance between source and target samples of the same label, $d_{W}^2(x_k,z_k) \leq u$ with $u$ as an upper bound, as well as thresholding the pairwise distance between source and target samples of different classes, $d_{W}^2(x_k,z_{k'}) \geq l$ with $l$ as a lower bound. This ensures that the learned metric regards samples of the same class but different domains as similar, while regarding samples of different classes as dissimilar. If no target labels are available, then one is required to encode similarity in other ways. 

ITML is restricted to finding transformations between domains in the same feature space. However, sometimes different descriptors are used for different image data sets. One descriptor might span a source feature space of dimensionality $D_{\mathcal{ S}}$ while another spans a feature space of dimensionality $D_{\mathcal{ T}}$. ITML is symmetric in the sense that it requires both domains to be of the same dimensionality. It is hence a domain adaptation technique, according to our definition in Section \ref{sec:domain}. But it can be extended to an asymmetric case, where $d_{W}(x,z) \neq d_{W}(z,x)$. Asymmetric Regularized Cross-domain transfer (ARC-t) incorporates non-square metric matrices $W^{D_{\mathcal{ S}} \times D_{\mathcal{ T}}}$ to find general mappings between feature spaces of different dimensionalities \cite{kulis2011you,hoffman2014asymmetric}. It is hence a transfer learning approach, according to our definitions. 

\paragraph{} Another means of estimating a cross-domain metric is to use the Fisher criterion. The Fisher criterion consists of the ratio of between-class scatter, $S_B = \sum_{k}^{K} \pi_k (\mu_k - \bar{\mu})(\mu_k - \bar{\mu})^{\top}$ with $\mu_k$ the mean of the $k$-th class, $\bar{\mu}$ the overall mean and $\pi_k$ the class-prior, and average within-class scatter, $S_W = \sum_{k}^{K} \pi_k \Sigma_{k}$ with $\Sigma_K$ as the covariance matrix of the $k$-th class \cite{fisher1938statistical,mclachlan2004discriminant}. The Fisher criterion can be used to extract a set of basis vectors that maintains class separability, much like a supervised form of PCA \cite{martinez2001pca}. FIDOS, a FIsher based feature extraction method for DOmain Shift \cite{dinh2013fidos}, is its extension to minimize domain separability while maintaining class separability. FIDOS incorporates multiple source domains and creates a between-class scatter matrix of the weighted average of the between-class scatter matrices in each domain as well as a within-class scatter matrix of the weighted average of each domains within-class scatter. It has a trade-off parameter to fine-tune the balance between the two objectives. 

Outside of estimating metrics for subsequent metric-based classifiers, there are also techniques that align class margins for subsequent maximum-margin classifiers \cite{hoffman2013efficient}.

\subsubsection{Manifolds} One can make an even stronger assumption than that of the existence of a mapping from source to target domain: that there exists a manifold of transformations between the source and target domain \cite{gong2013reshaping,baktashmotlagh2014domain,hoffman2014continuous,zhang2015multi}. A manifold is a curved lower-dimensional subspace embedded in a larger vector space. A transformation manifold corresponds to a space of parameters, where each point generates a possible domain. For example, it might consist of a set of camera optics parameters, where each setting would measure the data in one vector space basis \cite{gopalan2011domain,baktashmotlagh2014domain}. This assumption is useful when selecting a set of transformations as a geodesic path along the manifold \cite{gong2012geodesic,gong2013connecting,gopalan2014unsupervised,baktashmotlagh2013unsupervised,caseiro2015beyond}.

One of the first approaches to incorporate a transformation manifold looked at the idea of learning incremental small subspace transformations instead of a single large transformation \cite{gopalan2011domain}. Incremental learning was originally explored in situations with time-dependent concept shifts \cite{schlimmer1986incremental}. In our context, the goal is to learn the most likely intermediate subspaces between the source and target domain. The space of all possible $d$-dimensional subspaces in an $D$-dimensional vector space can be described by the \emph{Grassmann} manifold \cite{wong1967differential,absil2004riemannian,turaga2008statistical,zheng2012grassmann}. Each point on the Grassmannian generates a basis that forms a subspace \cite{gong2012geodesic,zheng2012grassmann,gong2013connecting}. One can move from one subspace to another by following the path along the surface of the manifold, also known as a geodesic. Computing the direction of the geodesic is performed using matrix exponentials or something called spline flow \cite{gallivan2003efficient,caseiro2015beyond}. Given the geodesic, all intermediate subspaces are computed and the source data is projected onto each of them separately. A classifier then trains on labeled data from a starting subspace and predicts labels for the next subspace, which are used as the labeled data in the next step. But the iteration can be replaced by integrating over the entire path. This produces a Geodesic Flow Kernel (GFK) which can be used in a nearest-neighbour classifier \cite{gong2012geodesic,gong2013connecting}. 

\paragraph{} Working with Grassmann manifolds for subspace mappings is just one option. Alternatively, one could look at statistical manifolds, where each point generates a probability distribution \cite{baktashmotlagh2014domain}. Geodesics along the statistical manifold describe how to transform one probability distribution into another. The length of the geodesic along the statistical manifold, called the Hellinger distance, can be used as a measure of domain discrepancy \cite{hellinger1909neue,ditzler2011hellinger}. The Hellinger distance is closely related to the total variation distance between two distributions, which is used in a number of other papers \cite{ben2007analysis,ben2010theory,mansour2014robust}. Adaptation consists of importance-weighting samples or transforming parameters to minimize the Hellinger distance \cite{baktashmotlagh2014domain,baktashmotlagh2016distribution}.

\subsection{Finding domain-invariant spaces}
The problem with transformations between domains is that the classifier remains in a domain-specific representation. Ideally, we would like to represent the data in a space which is domain-invariant.

\paragraph{} Most domain-invariant projection techniques stem from the computer vision and (biomedical) imaging communities, where domain adaptation and transfer learning problems are often caused by different acquisition methods. For instance, in computer vision, camera-specific variation between sets of photos is an unwanted factor of variation \cite{baktashmotlagh2013unsupervised,hoffman2013efficient}. In medical imaging, there exists a true representation of a patient and each MR scanner's image is a different noisy observation \cite{van2015transfer,kouw2017mr}. Since properties of the acquisition device are known, it is possible to design techniques specific to each type of instrument.

\paragraph{} Instead of finding principal components in each domain, one could also find common components. If the domains are similar in a few components, then both sets could be projected onto those components. In order to find common components where the sets are not too dissimilar, the Maximum Mean Discrepancy is employed \cite{pan2010cross,pan2011domain}. First, the MMD measure is rewritten as a joint domain kernel $\mathrm{K}$ \cite{pan2008transfer}. Now, a projection $C$ is introduced, which could be of lower dimensionality than the original number of features. From the joint domain kernel, components are extracted by minimizing the size of all projected data, where size corresponds to the \emph{trace} of the resulting matrix. The trivial solution to such a problem set up is that the projection maps everything to $0$. To avoid this, a constraint is added ensuring that the projection remains confined to a certain size. The resulting optimization problem becomes:
\begin{align}
	\underset{C}{\text{minimize}} \quad &\text{trace}(C^{\top}\mathrm{K} \mathrm{L} \mathrm{K} C) \nonumber \\
	\text{s.t.} \quad & C^{\top}\mathrm{K} \mathrm{H} \mathrm{K}C = I \, ,
\end{align}
where $\mathrm{L}$ the normalization matrix that divides each entry in the joint kernel by the sample size of the domain from which it originated, and $\mathrm{H}$ is the matrix that centers the joint kernel matrix $\mathrm{K}$ \cite{pan2011domain}. A regularization term $\text{trace}(C^{\top}C)$ along with a trade-off parameter, can be added as well. The solution to this optimization problem is an eigenvalue decomposition \cite{scholkopf1997kernel,scholkopf1998nonlinear}. 


\paragraph{} The Domain-Invariant Projection approach from \cite{baktashmotlagh2013unsupervised}, later renamed to Distribution Matching Embedding (DME) \cite{baktashmotlagh2016distribution}, aims to find a projection matrix $W$ that minimizes the MMD as follows:
\begin{align}
\hat{\mathrm{D}}^2_{\text{DME}}[w,X,Z] =& \ \frac{1}{n^{2}} \sum_{i,i'}^{n} \kappa(x_i W,x_{i'} W) - \frac{2}{mn} \sum_{i}^{n} \sum_{j}^{m} \kappa(x_i W,z_j) + C \, , \nonumber
\end{align}
where $C$ is a set of constant terms. The projection is constrained to remain orthonormal; $W^{\top}W = I$. Note that this formulation is similar to Kernel Mean Matching (c.f. Equation \ref{eq:kmm}), except that the projection operation is inside the kernel as opposed to the weighting function which is outside the kernel. 

Although the MMD encourages moments of distributions to be similar, it does not encourage smoothness in the new space. To this end, a regularization term can be added that punishes the within-class variance in the domain-invariant space. Futhermore, DME is still limited by its use of a linear projection matrix. A nonlinear projection is more flexible and more likely to recover a truly domain-invariant space. The Nonlinear Distribution-Matching Embedding achieves this by performing the linear projection in kernel space; $\phi(x)W$ \cite{baktashmotlagh2016distribution}. An alternative to finding a projection with the MMD is to use a Hellinger distance instead \cite{baktashmotlagh2016distribution}.

\paragraph{} Alternatively, MMD's kernel can be learned as well \cite{muandet2013domain}. Instead of using a universal kernel to measure discrepancy, it is also possible to find a basis function for which the two sets of distributions are as similar as possible. Considering that different distributions generate different means in kernel space, it is possible to describe a distribution of kernel means \cite{smola2007hilbert,berlinet2011reproducing}. The variance of this meta-distribution, termed \emph{distributional variance}, should then be minimized. However, this is fully unsupervised and could introduce class overlap. The functional relationship between the input and the classes can be preserved by incorporating a central subspace in which the input and the classes are conditionally independent \cite{gu2009learning,kim2011central}. Maintaining this central subspace during optimization ensures that classes remain separable in the new domain-invariant space. Overall, as this approach finds kernel components that minimize distributional-variance, it is coined Domain-Invariant Component Analysis (DICA) \cite{muandet2013domain}. It has been expanded on for the specific case of spectral kernels by \cite{long2015domain}.

\paragraph{} Several researchers have argued that in computer-vision settings there exists a specific lower-dimensional subspace that allows for maximally discriminating target samples based on source samples. Transfer Subspace Learning aims to find that subspace through minimizing a Bregman divergence to both domains \cite{si2010bregman}. The reconstruction error from the subspace to original space should be minimal \cite{shao2014generalized}. Interestingly, this objective is similar to that of an autoencoder, where the mapping is constructed through a neural network: $\| \phi(\phi(xW)W^{-1}) - z \|$ \cite{hinton1994autoencoders}. Deep autoencoders stack multiple layers of nonlinear functions to achieve flexible transformations \cite{vincent2010stacked}. They are mostly used in settings where large amounts of data are available \cite{glorot2011domain,chen2012marginalized}.

\paragraph{} Although the kernel approaches have the capacity to recover any nonlinear mapping, they require multiple computations of matrices as large as the number of samples. They therefore do not scale well. For larger data sets, one might employ neural networks. They also have the capacity to recover any nonlinear mapping but scale much better in terms of the number of samples \cite{bishop1995neural,rojas2013neural}. Neural networks are layered, and when going from one layer to the next, the input representation is transformed using a linear operation and pushed through a nonlinear activation function. By increasing the layer complexity and stacking multiple layers on top of each other, under certain conditions, any possible transformation can be achieved \cite{cybenko1989approximation,hornik1989multilayer,hornik1991approximation,bartlett1998sample}. Its optimization procedure, known as backpropagation, finds a transformation to a space in which the data is maximally linearly separable. By using different loss functions in the top layer, different forms of transformations can be achieved \cite{ganin2015unsupervised}. Domain-Adversarial Neural Networks (DANN) have one classification-error minimizing top layer and one domain-error maximizing top layer \cite{ganin2016domain}. Essentially, the network finds a domain-invariant space when its domain classifier cannot recognize from which domain a new sample has come. The idea of maximizing domain-confusion while minimizing classification error has inspired more approaches \cite{tzeng2015simultaneous,kamnitsas2017unsupervised}.



\subsection{Feature augmentation}
In natural language processing (NLP), text corpora show different word frequency statistics. People express themselves differently in different contexts \cite{blitzer2008domain,jiang2008domain}. For instance, the word 'useful' occurs more often to denote positive sentiment in kitchen appliance reviews than in book reviews \cite{blitzer2007biographies}. It can even happen that certain words occur \emph{only} in particular contexts, such as 'opioid receptors' in abstracts of biomedical papers but not in financial news \cite{blitzer2011domain}. Domains present a large problem for the field, as a document or sentence system cannot generalize across context.

\paragraph{} NLP systems can exploit the fact that words tend to signal each other; in a \emph{bag-of-words} (BoW) encoding each document is described by a vocabulary and a set of word counts \cite{sidorov2014syntactic}. Words that signal each other, tend to occur together in documents. Co-occurring words lead to correlating features in a BoW encoding. Correlating features can be exploited as follows: suppose a particular word is a strong indicator of positive or negative sentiment and only occurs in the source domain. Then one could find a correlating ''pivot'' word that occurs frequently in both domains. The word in the target domain that correlates most with the pivot word is said to correspond structurally to the original word in the source domain word. It could be a good indicator of positive versus negative sentiment as well \cite{blitzer2006domain}. Adaptation consists of augmenting the bag-of-words encoding with pivot words, learning correspondences and training on the augmented feature space \cite{blitzer2006domain,blitzer2011domain,chen2011co,li2014learning}. 

How to find corresponding features, or in general how to couple subspaces, is an open question. Note that with more features, there is a larger chance to find a good pivot feature. The earliest approaches have extracted pivot features through joint principal components or maximizing cross-correlation \cite{blitzer2007biographies,blitzer2011domain}. However, such techniques are linear and can only model linear relationships between features. Later approaches are more nonlinear through the use of kernelization or by employing "co-training" approaches \cite{chen2011co,li2014learning}.


\subsection{Minimax estimators}
When any of the aforementioned approaches are applied to settings where their assumptions are violated, then adaptation could be detrimental to performance \cite{provost2001robust,kouw2017target}. To ensure a robust level of performance, a worst-case setting can be assumed. Worst-case settings are often formalized as minimax optimization problems \cite{berger2013statistical,grunwald2004game}. The variable that we want to be robust against is maximized before the risk is minimized with respect to classifier parameters. Minimax estimators behave conservatively and are not as powerful as standard estimators when their assumptions are valid. 

\paragraph{} A straightforward example of a minimax estimator is the Robust Bias-Aware classifier \cite{liu2014robust}. In this case, the uncertain quantity corresponds to the target domain's posterior distribution. However, given full freedom, the adversary would flip the labels of the classifier in every round to maximize the classifier's risk and convergence would not be achieved. A constraint is imposed, telling the adversary that it needs to pick posteriors that match the moments of the source distribution's feature statistics:
\begin{align}
	\underset{h \in \mathcal{H}}{\min} \ \underset{g \in \mathcal{H}}{\max} \quad &\frac{1}{m} \sum_{j=1}^{m} \ell( h(z_j), g(z_j))  \\ 
	\text{s.t.} \quad &\frac{1}{m} \sum_{j=1}^{m} g(z_j) = \frac{1}{n} \sum_{i=1}^{n} y_i \, , \nonumber
\end{align}	
where the constraint states that the first-order moment (i.e. the sample average) of the adversary's posteriors should be equal to first-order moment of the source label distribution. Higher-order moments can be included as constraints as well. This minimax estimator returns high confidence predictions in regions with high probability mass under the source distribution and uniform class predictions in regions with low source probability mass.

\paragraph{} Importance-weighted classifiers are sensitive to poorly estimated weights. To construct a robust version, one could minimize risk with respect to worst-case weights \cite{wen2014robust}:
\begin{align}
	\underset{h \in \mathcal{H}}{\min} \ \underset{w \in W}{\max} \quad &\frac{1}{n} \sum_{i=1}^{n} \ell( h(x_i), y_i) w_i \nonumber \\
		\text{s.t.} \quad& w_i \geq 0 \nonumber \\
	\quad& | \  \frac{1}{n} \sum_{i}^{n} w_i - 1 \ | \leq \epsilon \, .	\nonumber
\end{align} 
The weights will have to be constrained as otherwise they would tend to infinity. These constraints match those of the non-parametric weight estimators (c.f. Equation \ref{eq:nonparam_iw}). By training a classifier under worst-case weights, it will behave more conservatively. 

\subsection{Robust algorithms} \label{sec:robust}
Conservatism can also be expressed through \emph{algorithmic robustness} \cite{xu2012robustness}. An algorithm is deemed robust if it can separate a labeled feature space into disjoint sets such that the variation of the classifier is bounded by a factor dependent on the training set. Intuitively, a robust classification algorithm does not change its predictions much whenever a training sample is changed.

\paragraph{} This notion can be employed to construct a robust adaptive algorithm \cite{mansour2014robust}. Such an algorithm will try to find a separating hyperplane such that the hinge loss on the source set is similar to the loss on the target set. Essentially, the classifier is discouraged from picking support vectors in areas with low probability mass under the target distribution. The downside of this approach is that if the class posterior distributions of both domains are very different (e.g. orthogonal decision boundaries), it will not perform well on \emph{both} sets.

\section{Discussion} \label{sec:discussion}
A number of points come to mind, as well as ideas for future work.

\subsection{Validity of the covariate shift assumption}
The current assumption in covariate shift, namely $p_{\mathcal{ T}}(y \given x) = p_{\mathcal{ S}}(y \given x)$, might be too restrictive to ever be valid in nature. The assumption is often interpreted as that the decision boundary should be in the same location in both domains, but that is false. The posterior distributions need to be equal for the \emph{whole} feature space. Equal class-posterior distributions is a much more difficult condition to satisfy than equal decision boundaries. As such, there are many occasions where the assumption is made, but is not actually valid.

\paragraph{} Fortunately, some experiments have indicated that there is some robustness to a violation of the covariate shift assumption. It would be interesting to perform a perturbation analysis to study a classifier's behaviour as a function of the deviation from equal posteriors. Perhaps decision boundaries that lie within a specified $\epsilon$ distance from each other, are not too far apart to change a classifier's behaviour. That would mean importance-weighting is more widely applicable than its assumption implies. A less restrictive condition would be easier to satisfy, and methods that depend on it would be less at risk of negative transfer.

\subsection{Application-specific discrepancies}
Most measures that describe discrepancies between domains are general; they are either distribution-free or classifier-agnostic. General measures produce looser generalization bounds than more specific measures. As new insights are gained into causes of domain shift, more precise metrics should be developed. These can contain prior knowledge on the problem at hand: for example, in natural language processing, one often encodes text documents in bag-of-word or n-gram features. General measures such as the Maximum Mean Discrepancy might show small values for essentially entirely different contexts. A more specific measure, such as the total variation distance between Poisson distributions, would take the discreteness and sparseness of the feature space into account. Consequently, it would be more descriptive and it would be preferable for natural language processing domains. Such specific forms of domain discrepancy metrics would lead to tighter generalization bounds, stronger guarantees on classifier performance and more practical adaptive classifiers.

\paragraph{} Finding domain discrepancies specific to a task or type of data is not a trivial task. A good place to start is to look at methods that incorporate explicit descriptions of their adaptations. For instance, a subspace mapping method explicitly describes what makes the two domains different (e.g. lighting or background). Looking at the types of adaptations they recover would be informative as to what types of discrepancies are useful for specific  applications. Methods with explicit descriptions of ''transfer'' could be exploited for finding more specific domain discrepancy metrics.

\subsection{Open access and institution-variation}
It is not uncommon to hear that different research groups working in the same field do not use each other's data. The argument is that the other group is located in a different environment, experiments differently or uses a different measuring device, and that their data is therefore not ''suitable'' \cite{leek2010tackling}. For example, in biostatistics, gene expression micro-array data can exhibit ''batch effects'' \cite{johnson2007adjusting}. These can be caused by the amplification reagent, time of day, or even atmospheric ozone level \cite{fare2003effects}. In some data sets, batch effects are the most dominant source of variation and are easily identified by clustering algorithms \cite{johnson2007adjusting}. However, additional information such as local weather, laboratory conditions, or experimental protocol, should be available. That information could be exploited to correct for the batch effect. The more knowledge of possible confounding variables, the better the ability to model the transfer from one batch to the other. Considering the financial costs of genome sequencing experiments, the ability to combine data sets from multiple research centers without batch effects is desirable.

\paragraph{} This can be taken a step further: being able to classify a data set by downloading a source domain and training a domain-adaptive classifier instead of annotating samples, will save everyone time, funds and effort. Not only does it increase the value of existing data sets, it can also increase statistical power by providing more data. The development of domain-adaptive or transfer learning methods creates a larger incentive for researchers to make their data publicly available.

\subsection{Sequential adaptation}
"Successful" adaptation is defined as an improvement over the performance of the original system, i.e. a positive transfer. But settings where the domains are too dissimilar are too difficult. The more similar the populations are, the likelier it is that the system adapts well. But that raises the question: can a system be designed that first adapts to an intermediate population and only then adapts to the final target population? In other words, a system that \emph{sequentially} adapts?

\paragraph{} When incorporated, data from intermediate domains would present a series of changes instead of one large jump. For example, adapting from European hospitals to predicting illnesses in Asian hospitals is difficult. But a sequential adaptive system starting in western Europe would first adapt to eastern Europe, followed by the Middle-East, then to west Asia and finally reaching a population of eastern Asian patients. If the domain shifts are not too dissimilar in each transition, then adaptation should be easier. 

Note that this differs from the domain manifolds approach in Section \ref{sec:submappings}, in that those assume that there exists a space of transformations mapping source to target and everything in between. Here, that is not necessary; intermediate domain data could assist any type of method, not just mappings. For instance, it would be possible to perform a series of importance-weighting steps. Such an approach overlaps with \emph{particle filtering} somewhat \cite{doucet2000rao,djuric2003particle}. Particle filters weight the signals in each time-step with respect to the signals in the next time-step. Considering the similarity to importance-weighing in covariate shift, particle filters could be directly applicable to sequential domain adaptation.

\paragraph{} But the sequential strategy also raises a number of extra questions: will adaptation errors accumulate? How should performance gain be traded off against additional computational cost? Will performance feedback be necessary? Some of these questions have been addressed in dynamical learning settings, such as reinforcement learning or multi-armed bandits \cite{sutton1998reinforcement,whittle1980multi}. The design of sequential adaptive systems could build upon their findings, but more work shall need to be done.

\section{Conclusion} \label{sec:conclusion}
We posed the question: when and how can a classifier generalize from a source to a target domain? In the most general case, where there are no constraints on the difference between the source and target domain, no guarantees of performance above chance level can be given. For particular cases, certain guarantees can be made, but these still depend on the domain dissimilarity. Such dependencies lead to many new questions.



\paragraph{} Covariate shift is perhaps the most constrained case. In its simplest form, only one covariate / feature changes. This has been extensively studied and many questions, such as how the generalization error depends on the distance between domains, have been addressed. It seems that the most important thing to check before attempting a method is: does the assumption of equal class-posterior distributions hold and if not, how strongly is it violated? 

Following that, one could select an importance weight estimator based on the following questions: are the domains so far apart that the weights will become bimodal? Do you have enough source and target samples for complex estimators? If weight estimation is done parametrically, do you have enough samples to prevent low probabilities in the denominator. If done non-parametrically, do you have enough samples to perform hyperparameter estimation? 

But a number of additional questions remain: how does data preprocessing affect importance weight estimation versus classifier training? Should each domain be normalized separately to bring the domains closer together thereby avoiding weight bimodality or should this be avoided because it induces a violation of the covariate shift assumption? Does the assumption of equal class-posterior distributions hold for only a part of feature space?


\paragraph{} General cases of domain shift are more complicated. Multiple factors change at the same time and cannot be always be uniquely identified. For example, if there are both changes in the data distributions and changes in the class-posterior distributions, then these cannot be identified without \emph{some} target labels. Since domain shifts are less constrained, they are harder to study. It is thus difficult to predict whether a specific adaptive method will be successful for a given problem. Furthermore, it is still unclear what the effects of sample sizes or estimation errors are for methods based on subspace mappings, domain-invariant spaces, domain manifolds, low error of the ideal joint hypothesis, etc. It would be informative to study these factors.

\paragraph{} In conclusion, we would argue that there is still a lot to be done before transfer learners and domain-adaptive classifiers become practical tools.

\section*{Acknowledgements}
The author would like thank Marcel Reinders, Tom Viering and Soufiane Mouragui for feedback and discussions.

\newpage
\bibliography{kouw_techreport18a}
\bibliographystyle{plain}

\end{document}